\documentclass{article}

    \PassOptionsToPackage{numbers, compress}{natbib}

    \usepackage[preprint]{neurips_2024}

\usepackage[utf8]{inputenc} 
\usepackage[T1]{fontenc}    
\usepackage{hyperref}       
\hypersetup{
	colorlinks=true,
	linkcolor=red,
	filecolor=blue,      
	urlcolor=magenta,
	citecolor=black,
}
\usepackage{url}            
\usepackage{booktabs}       
\usepackage{amsfonts}       
\usepackage{nicefrac}       
\usepackage{microtype}      
\usepackage{xcolor}         
\usepackage{multirow}
\usepackage{graphicx}
\usepackage{amsmath}
\usepackage{amssymb}
\usepackage{cleveref}
\usepackage{xspace}
\usepackage{ulem}
\newcommand{\method}{\textsc{EditWorld}\xspace}
\newcommand{\zbh}[1]{\textcolor{black}{#1}}

\title{\method: Simulating World Dynamics for Instruction-Following Image Editing}

\author{%
  Ling Yang{$^{1}$}\thanks{These authors contributed equally.} \ \thanks{Correspondence to: yangling0818@163.com, wentao.zhang@pku.edu.cn.}, \ Bohan Zeng{$^{1}$}\footnotemark[1], \ Jiaming Liu{$^{2}$}, \ Hong Li{$^{1}$}, \\
  \textbf{Minghao Xu$^{4}$, \ Wentao Zhang{$^{1 \dag}$}, \ Shuicheng Yan$^{3}$} \\
  {$^{1}$} Peking University, {$^{2}$} Tiamat AI, {$^{3}$} Skywork AI, {$^{4}$} Mila - Québec AI Institute
}

\begin{document}

\maketitle

\begin{abstract}
  Diffusion models have significantly improved the performance of image editing. Existing methods realize various approaches to achieve high-quality image editing, including but not limited to text control, dragging operation, and mask-and-inpainting. Among these, instruction-based editing stands out for its convenience and effectiveness in following human instructions across diverse scenarios. However, it still focuses on simple editing operations like adding, replacing, or deleting, and falls short of understanding aspects of \textit{world dynamics} that convey the realistic dynamic nature in the physical world. Therefore, this work, \method, introduces a new editing task, namely \textit{world-instructed image editing}, which defines and categorizes the instructions grounded by various world scenarios. We curate a new image editing dataset with world instructions using a set of large pretrained models (e.g., GPT-3.5, Video-LLava and SDXL). To enable sufficient simulation of world dynamics for image editing, our \method trains model in the curated dataset, and improves instruction-following ability with designed \textit{post-edit} strategy. Extensive experiments demonstrate our method significantly outperforms existing editing methods in this new task. Our dataset and code will be available at \href{https://github.com/YangLing0818/EditWorld}{https://github.com/YangLing0818/EditWorld}
\end{abstract}

\section{Introduction}

Text-to-image models have achieved great success with the significant development of diffusion models \cite{song2020score,ho2020denoising,yang2023diffusion}, such as Stable Diffusion \cite{rombach2022high}, DALL-E 2/3 \cite{ramesh2022hierarchical,betker2023improving}, and RPG \cite{yang2024mastering}.
In order to improve controllability, text-guided image editing has gained remarkable improvements \cite{crowson2022vqgan,gal2022stylegan,hertz2022prompt,yang2024mastering}. Early researches such as StyleGAN \cite{gal2022stylegan} and CycleGAN \cite{zhu2017unpaired} are able to achieve reasonable image style transfer. Subsequent models \cite{sohn2023styledrop,zhang2024real,ye2023ip,zhang2024ssr,nam2024dreammatcher} mainly extract key features from images, which can be further incorporated into the text-to-image generation process to generate images with the style and content of reference images. 

Aiming to make visual editing more precise, some methods try to enable dragging operations \cite{pan2023drag,shi2023dragdiffusion,mou2024dragondiffusion} or condition on additional masks \cite{avrahami2023blended,ye2023ip,zhang2024ssr} for localized image manipulation.  
Thanks to the quick development of multimodal pretraining methods like CLIP models \cite{radford2021learning}, numerous methods \cite{bar2022text2live,mokady2022null,tumanyan2023plug,couairon2022diffedit,yang2024crossmodal} leverage expressive cross-modal semantic controls to guide the text-based image editing process without additional manual operations or masks.  

Recently, instruction-based image editing \cite{brooks2023instructpix2pix,zhang2023magicbrush,li2023zone,huang2023smartedit,fu2024guiding} teaches the image editing model to follow human instructions, allowing ordinary users to conveniently and effortlessly manipulate images through natural commands. For instance, InstructPix2Pix \cite{brooks2023instructpix2pix} leverages two large pretrained models (i.e., GPT-3 and Stable Diffusion) to generate a large dataset
of input-goal-instruction triplet examples, and trains an instruction-following image editing model on the dataset. MGIE \cite{fu2024guiding} further incorporates multimodal large language models (MLLMs) to facilitate editing instructions by providing more expressive visual-aware instructions.  

\begin{figure}[t]
\vspace{-6mm}
  \centering
  \includegraphics[width=0.98\linewidth]{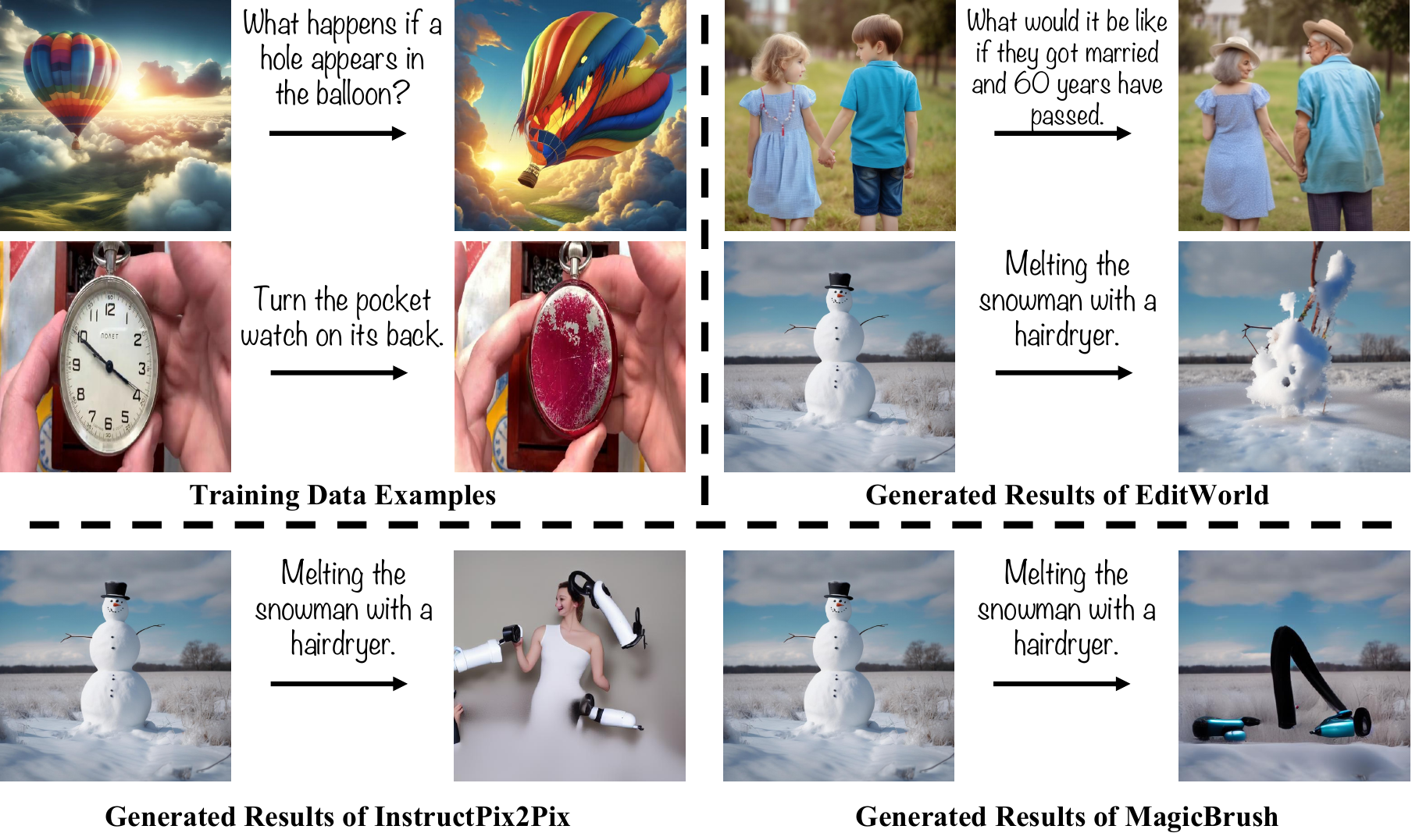}
  \caption{\method offers 10k sets of input and output images with instructions for  world-instructed image editing task. As shown in the figure, our \method performs better than both InstructPix2Pix \cite{brooks2023instructpix2pix} and MagicBrush \cite{zhang2023magicbrush} in image editing involving world dynamics.}
  \label{fig:first_fig}
  \vspace{-5mm}
\end{figure}

Despite the high-quality results achieved by existing text- and instruction-based methods, they ignore and also struggle to deal with the \textit{world dynamics} that convey the realistic visual dynamic of the physical world \cite{ha2018world} in image editing. For example, they can achieve reasonable results when instructed to "put a hat on the man", but they may fail to edit with the instruction "what happens if the man slips". As shown in Fig.~\ref{fig:first_fig}, neither InstructPix2pix \cite{brooks2023instructpix2pix} nor MagicBrush \cite{zhang2023magicbrush} can generate reasonable editing results.

To enable image editing to reflect complex \textit{world dynamics} from both real physical world and virtual media, this work \method, introduces a new task named \textit{world-instructed image editing}, as the data examples presented in Fig.~\ref{fig:first_fig}. Specifically, we define and categorize diverse world instructions, and based on them curate a new multimodal dataset that consists of a large number of input-instruction-output triplets. 
The construction of this dataset has two branches: (1) We utilize GPT-3.5 to generate world instructions for different scenarios and employ text-to-image diffusion models such as SDXL \cite{podell2024sdxl} and ControlNet \cite{zhang2023adding} to produce high-quality input-output image pairs that strictly follow the world instructions; (2) We select two frames from video data as paired image data, which contains consistent identity between consecutive frames but has large spatial variances or dynamics caused by  video motions. Then we utilize Video-LLava \cite{lin2023video} and GPT-3.5 to generate corresponding instructions.

Finally, we finetune a instructPix2Pix model using our curated dataset and propose a zero-shot image manipulation strategy \textit{post-edit} to improve the ability of instruction following and enhance the appearance consistency of non-editing areas. Extensive experiments demonstrate our method achieve state-of-the-art (SOTA) results in the world-instructed image editing task compared to previous methods. We summarize our main contributions as follows:
\begin{itemize}
    \item We propose a new fundamental task named \textit{world-instructed image editing}, which reflects the real dynamics in both physical world and virtual media. We define and categorize world instructions according to different scenarios.

    \item We curate a new multimodal dataset for world-instructed image editing by leveraging a set of large pretrained models, including GPTs, MLLMs, and text-to-image diffusion models. This curated dataset can also serve as a new benchmark for our proposed editing task.

    \item We train and improve our diffusion-based image editing model on the dataset, outperforming previous state-of-the-art (SOTA) image editing methods in context of world-instructed image editing while maintaining competitive results in traditional editing tasks.
    
\end{itemize}


\section{Related Works and Discussions}

\subsection{Diffusion Models for Text-Guided Image Editing}
Recent advancements in Diffusion models (DMs) \cite{ho2020denoising, song2020denoising, dhariwal2021diffusion, vahdat2021score, ho2022cascaded, rombach2022high, peebles2022scalable,yang2024mastering,yang2024crossmodal,yang2024structure,zhang2024realcompo,yang2023improving} and multimodal frameworks such as CLIP \cite{radford2021learning} have significantly enhanced text-to-image generation, and demonstrate impressive generation results \cite{huang2024diffusion, brooks2023instructpix2pix, kawar2023imagic, yu2022scaling, meng2021sdedit, ramesh2022hierarchical, saharia2022photorealistic, nichol2022glide,ruiz2023dreambooth,zeng2024controllable, gao2023implicit}. Building upon these foundations, some works \cite{kim2022diffusionclip, kwon2022clipstyler, avrahami2022blended, avrahami2023blended, hertz2022prompt, parmar2023zero, tumanyan2023plug} realize text-guided image editing using conditional diffusion models \cite{ho2022classifier}. For example, Blend Diffusion \cite{avrahami2022blended} introduces a mask-guided editing method for precise and seamless image manipulation.
Prompt-to-Prompt \cite{hertz2022prompt} manipulates an image to align with the words in the prompt by modifying cross-attention, allowing for localized image editing through changing textual prompts. 
PAIR-diffusion \cite{goel2023pairdiffusion} provides a way to independently edit both the structure and appearance within each masked area of the input image. Although these methods perform impressively, they highly depend on elaborate textual descriptions or precise regional masks. In contrast, providing direct instructions to modify specific regions or attributes, such as 'make the tie blue', offers a more straightforward and user-friendly way for ordinary users.

\subsection{Instruction-Following Image Editing}
Instruction-based image editing \cite{el2019tell, zhang2021text, ouyang2022training, zhang2023hive, brooks2023instructpix2pix, zhang2023magicbrush,li2023zone} aims to teach a generative model to
follow human-written instructions for image editing, and has made some progress. For instance, InstructPix2Pix \cite{brooks2023instructpix2pix} employs a large language model GPT-3 \cite{brown2020language} and Prompt-to-Prompt \cite{hertz2022prompt} to synthesize a training dataset for obtaining a diffusion model specialized in instruction-following image editing. MagicBrush \cite{zhang2023magicbrush} enhances the capabilities of InstructPix2Pix by fine-tuning it with a real collected image dataset.
Another line of researches utilizes multimodal large language models (MLLMs) to enhance the cross-modal understanding for instruction-based image editing \cite{fu2024guiding, huang2023smartedit}. For example, MGIE \cite{fu2024guiding} jointly learns the MLLM and editing model with vision-aware expressive instructions to provide explicit guidance.
Although existing image editing methods, especially instruction-based image editing methods, can already make simple editing operations in images, they still struggle to deal with the instructions from real physical world and also lack a dataset that contains image pairs reflecting world dynamics. To address this issue, we formulate a new task setting, namely \textit{world-instructed image editing}, and achieve SOTA performance under this setting.

\subsection{Leveraging Large Pretrained Models for Multimodal Task}
With the development of large-scale pretraining techniques, many works leverage the extensive knowledge embedded in pretrained large models (e.g., (M)LLM, SAM or Stable Diffusion) \cite{kirillov2023segment,rombach2022high} to more effectively addressing various tasks. Notable achievements have been made in some areas, such as image editing \cite{hertz2022prompt, li2023zone, fu2024guiding, brooks2023instructpix2pix, zhang2023magicbrush}, video editing \cite{wu2023tune, geyer2023tokenflow, zeng2023face} and text-to-3D generation \cite{poole2022dreamfusion, lin2023magic3d, zeng2023ipdreamer}. Our \method leverages a set of large pretrained models to process both image and video data, and construct a multimodal dataset that contains diverse input-instruction-output triplet examples for enhancing world-instructed image editing. 

\subsection{Data Generation with Generative Models}
Deep models generally need vast amounts of data for training. While internet offers a rich data resource, it often lacks the structured, paired data needed for supervised learning. As generative models advance, they are increasingly seen as a cost-effective way to produce large quantities of training data for various applications \cite{li2022bigdatasetgan, peebles2022gan, tritrong2021repurposing, viazovetskyi2020stylegan2, brooks2023instructpix2pix, zhang2023magicbrush, jin2024better,jin2024mm}. The paired data required for our proposed task setting is rarely found on the internet. In this work, we design two synthesis branches, i.e., text-to-image generation and video frame extraction, and utilize a combination of generative models to generate complex and rich data samples for world-instructed image editing.

\section{Method}

\subsection{Definition of World Instructions}

Instruction-guided image editing aims to edit the input images according to the instruction. Existing methods \cite{brooks2023instructpix2pix,zhang2023magicbrush} have effectively achieved object replacing, removing, and adding. However, it remains a challenge for existing methods to simulate the real visual dynamics in physical world or virtual media. 
Therefore, in this section, we will introduce how we generate a multimodal dataset containing editing triples that can reflect various world dynamics.
We first define and categorize our proposed world instructions. 
The images we encounter typically originate from either physical world or virtual media. Consequently, we categorize world instructions into two main types: real-world and virtual-world. We further refine the two types of world instructions into seven specific categories based on the different scenarios, as shown in Table~\ref{data_example}.
To enhance the comprehension of different world instructions, we provide an example input-instruction-output triplet to instantiate each type.

\begin{table}[t]
  \caption{Definition and examples of different world instructions in our generated dataset.}
  \label{data_example}
  \centering
  \renewcommand{\arraystretch}{1.5}
  \small
  \begin{tabular}{lp{4cm}lp{2.cm}l}
    \toprule
    Category & Description & Input & Instruction & Output \\ \hline
    \multicolumn{5}{c}{\textbf{Real-world Instruction}} \\ \hline
    \textit{Long-Term}      & In this editing type, there is a significant time interval between the content of the input and the output images. &  \raisebox{-0.8\totalheight}{\includegraphics[width=0.14\linewidth]{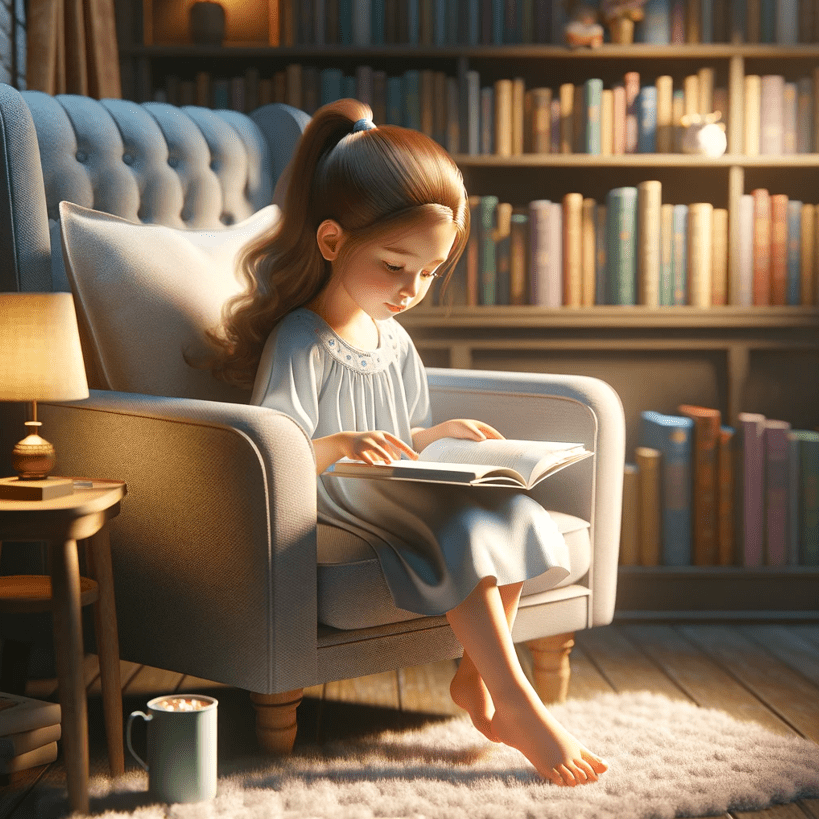}}      &  ``What would have happened if the girl had kept on reading and writing?''  & \raisebox{-0.8\totalheight}{\includegraphics[width=0.14\linewidth]{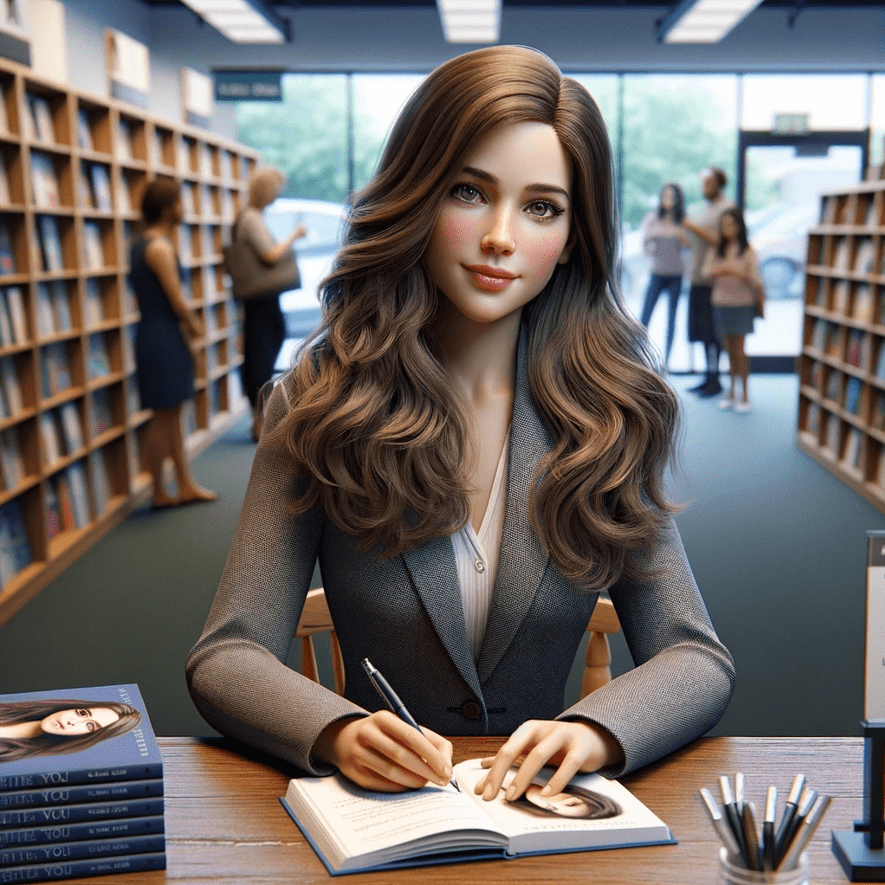}}   \\
    \textit{Spatial-Trans}     & In this editing type, the position of the object in the image or the viewpoint of the image undergoes substantial shifts or transformations. & \raisebox{-0.8\totalheight}{\includegraphics[width=0.14\linewidth]{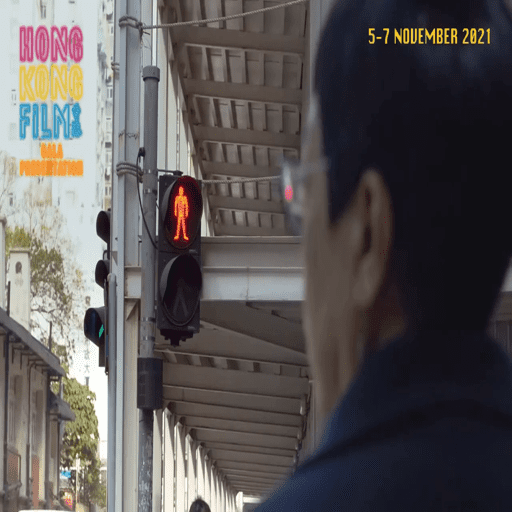}} & ``The traffic light becomes passable.'' &  \raisebox{-0.8\totalheight}{\includegraphics[width=0.14\linewidth]{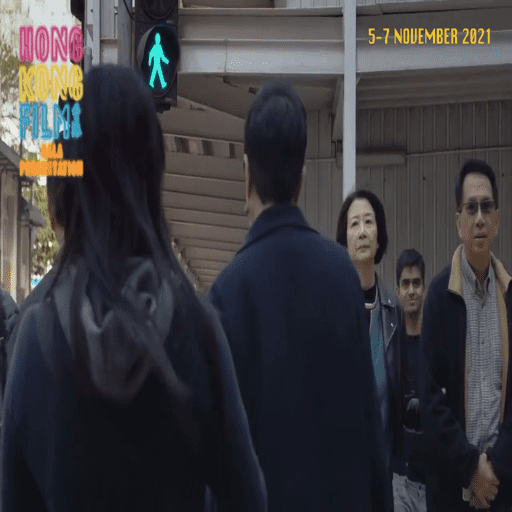}}  \\ 
    \textit{Physical-Trans}     & In this editing type, the physical characteristics of objects in the image, including shape, structure, and texture, exhibit significant changes.  & \raisebox{-0.8\totalheight}{\includegraphics[width=0.14\linewidth]{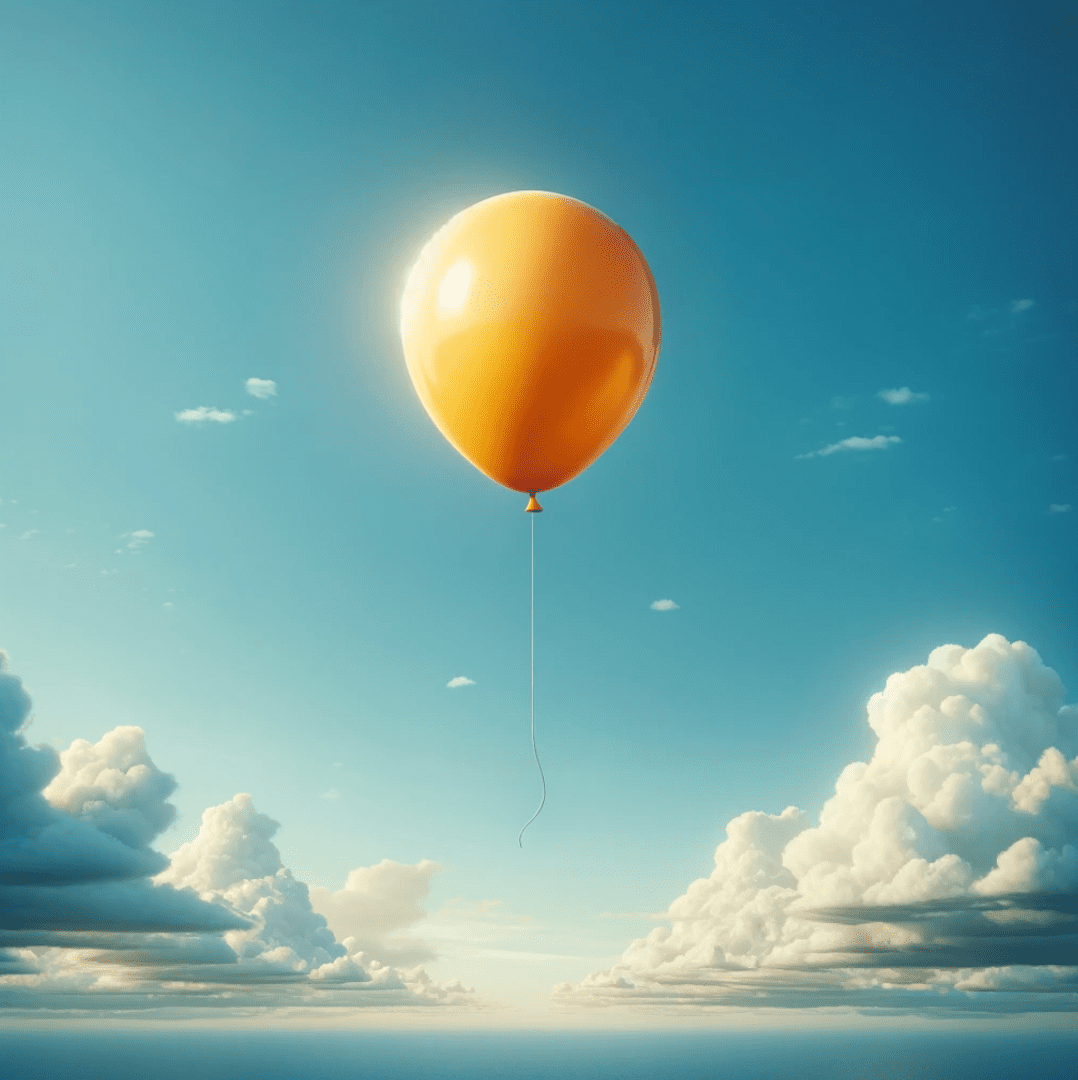}}  &  ``Popping the balloon.'' & \raisebox{-0.8\totalheight}{\includegraphics[width=0.14\linewidth]{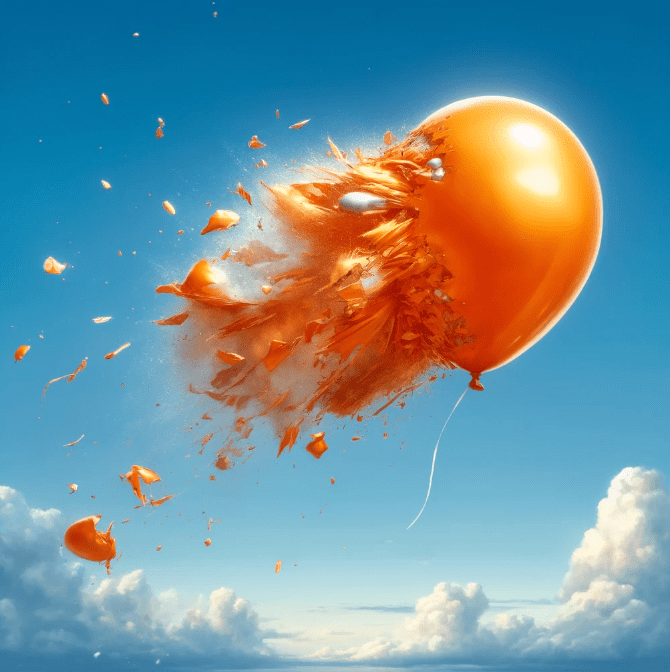}}       \\
    \textit{Implicit-Logic}     & In this editing type, the editing instructions often imply very implicit logic.  & \raisebox{-0.8\totalheight}{\includegraphics[width=0.14\linewidth]{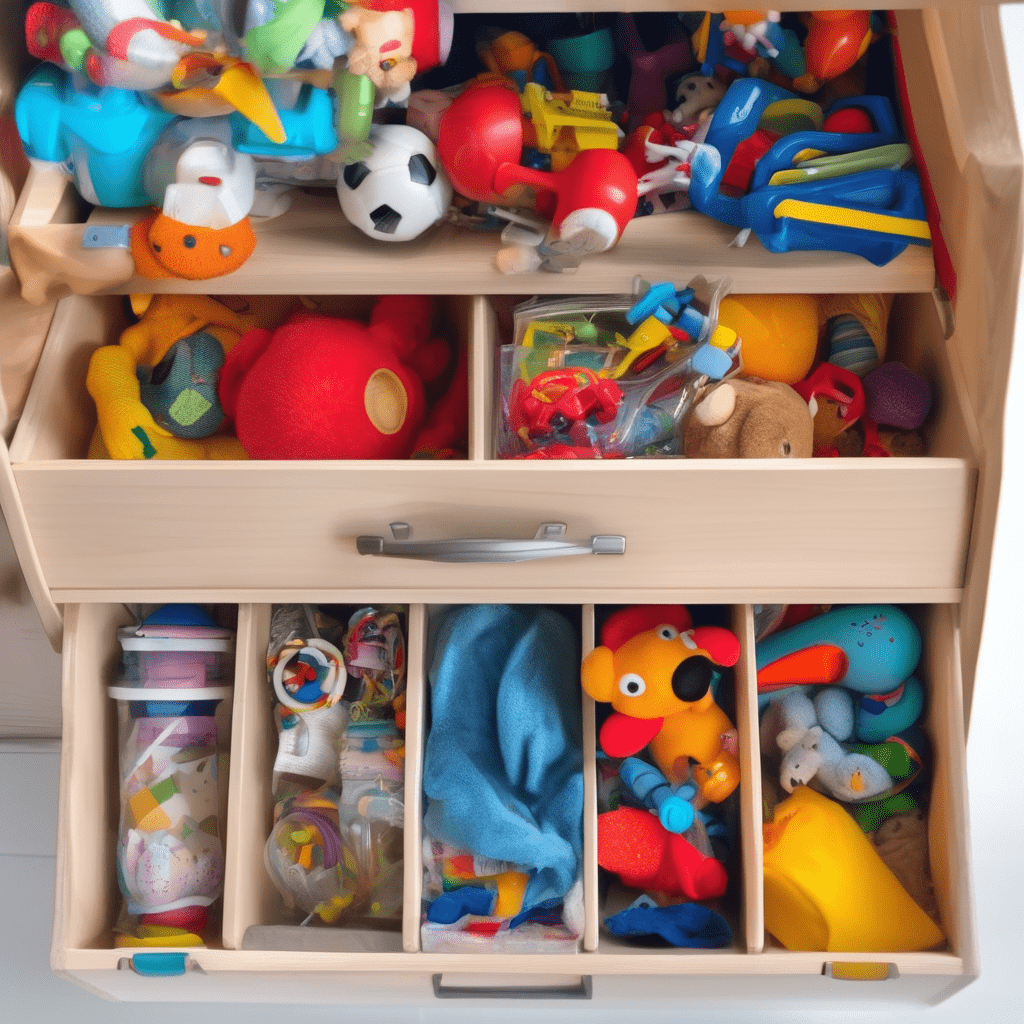}}  & ``what would happen if a impatient child look for a specific toy?'' & \raisebox{-0.8\totalheight}{\includegraphics[width=0.14\linewidth]{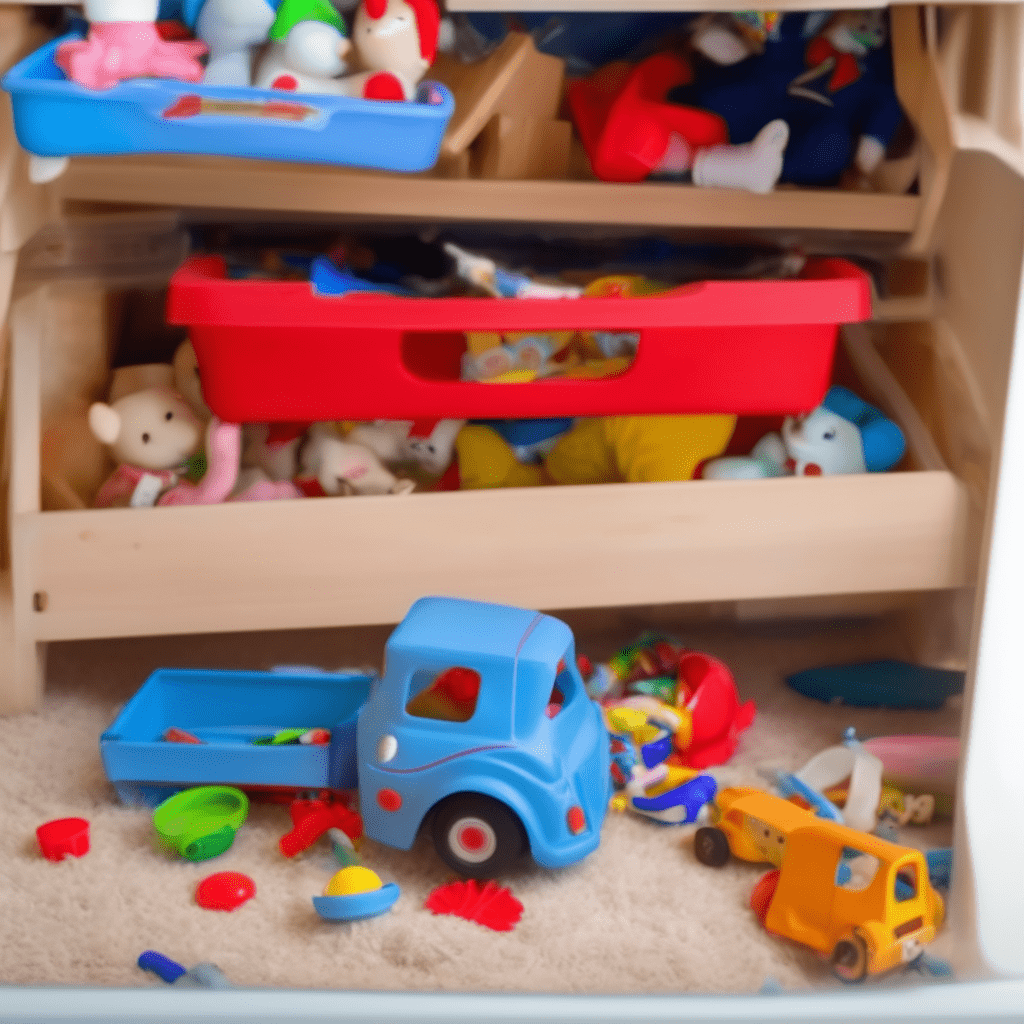}}       \\ \hline
    \multicolumn{5}{c}{\textbf{Virtual-world Instruction}} \\ \hline
    \textit{Story-Type}     & In this editing type, the edits often relate to the storyline of the fairy tale or movie.  & \raisebox{-0.8\totalheight}{\includegraphics[width=0.14\linewidth]{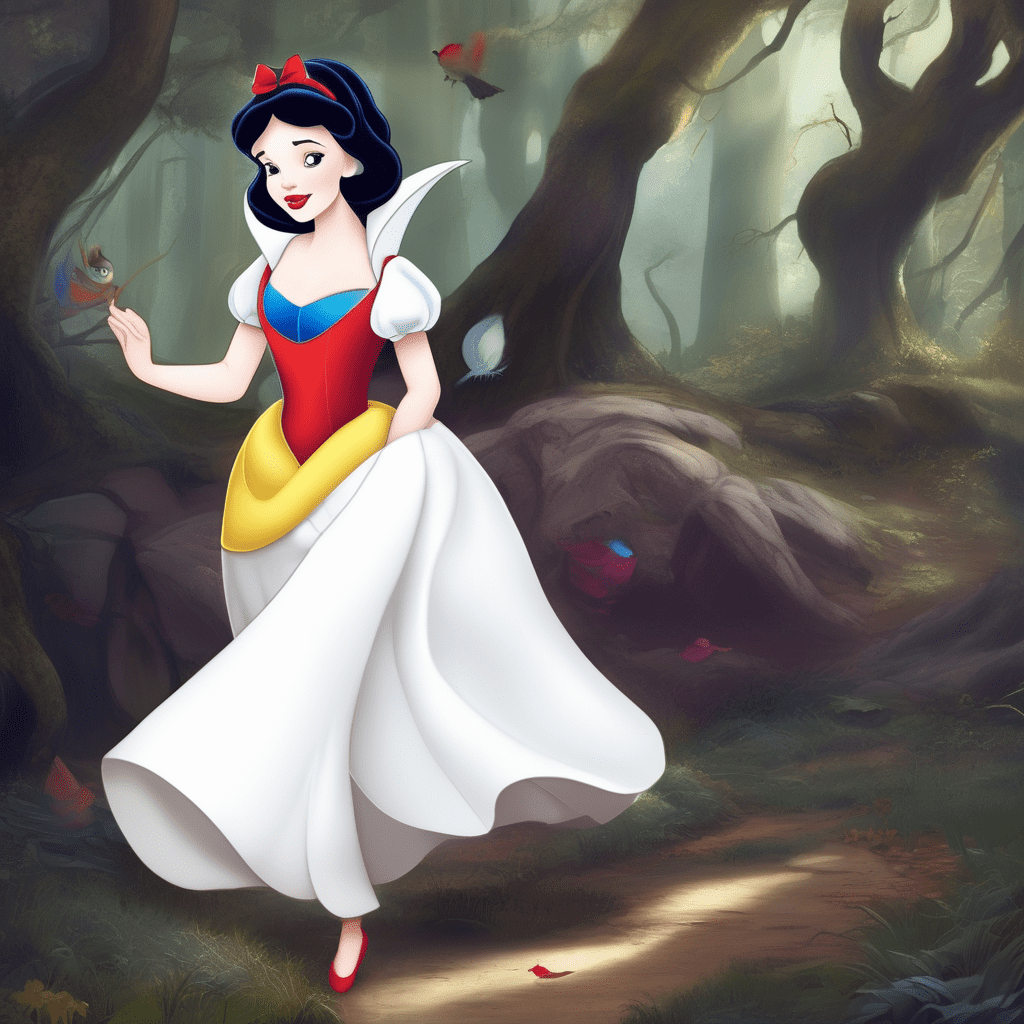}} & ``what would happen if Snow White ate a poisoned apple?'' & \raisebox{-0.8\totalheight}{\includegraphics[width=0.14\linewidth]{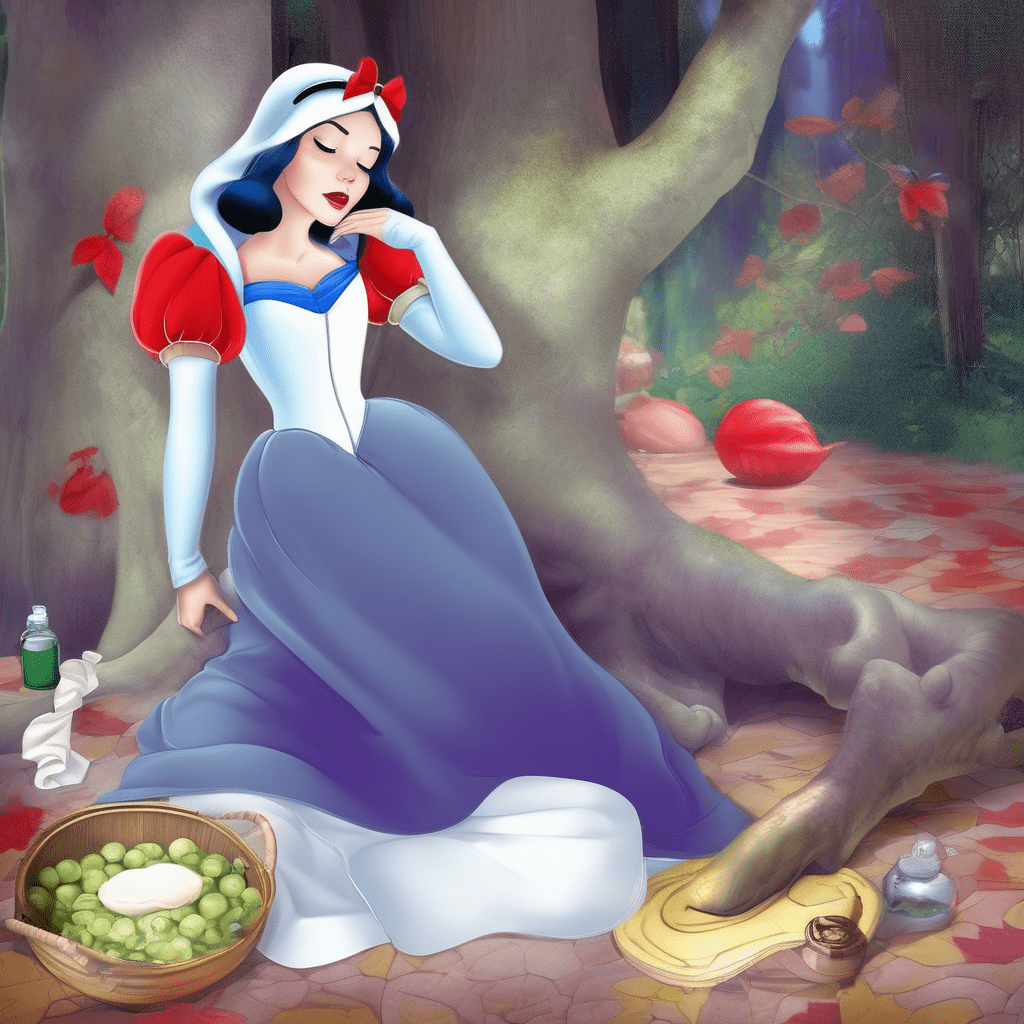}}       \\
    \textit{Real-to-Virtual}     & In this editing type, natural phenomena from the real world are introduced into specific virtual world scenarios.  & \raisebox{-0.8\totalheight}{\includegraphics[width=0.14\linewidth]{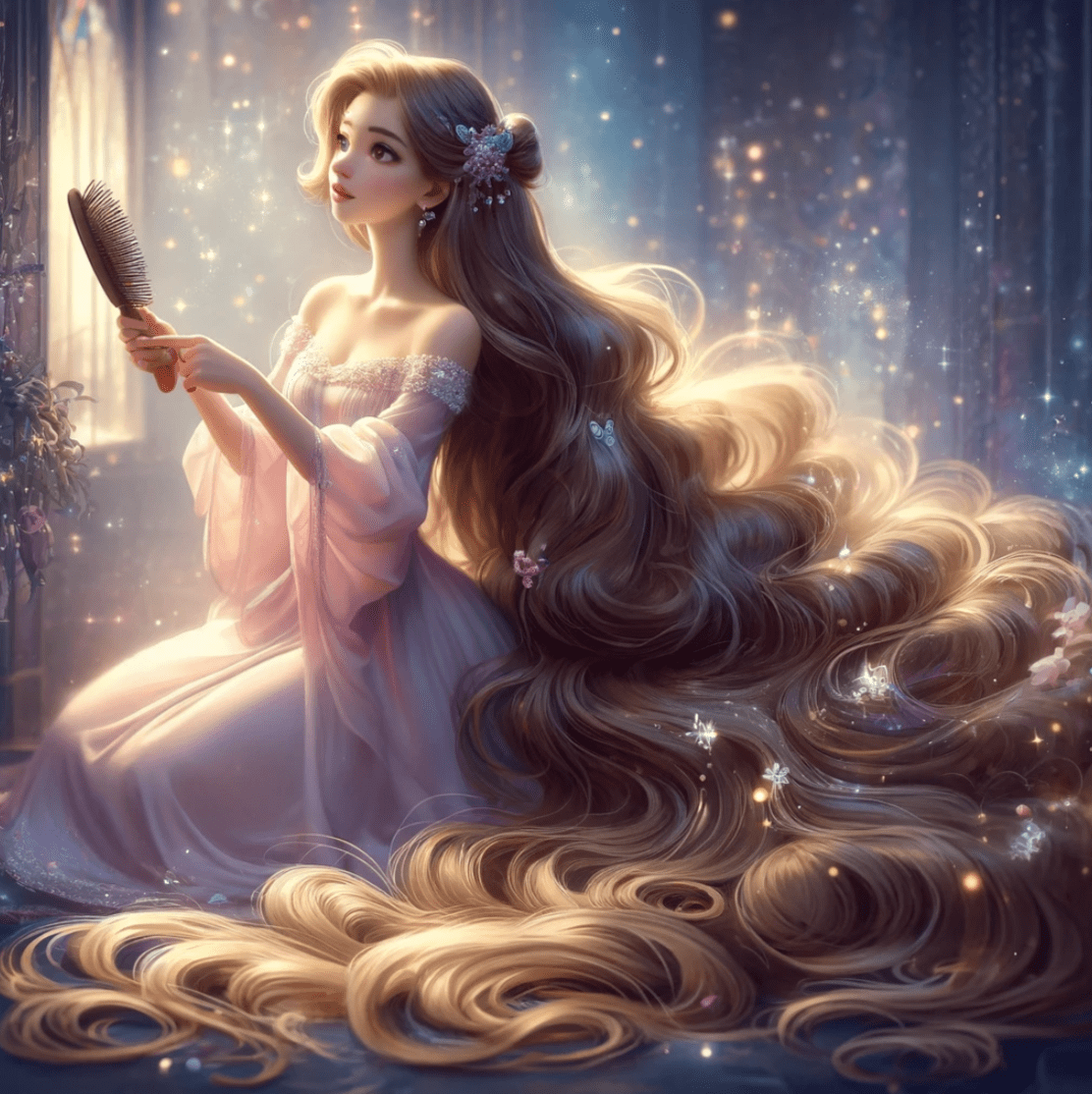}} & ``what would happen if there was strong static electricity?'' & \raisebox{-0.8\totalheight}{\includegraphics[width=0.14\linewidth]{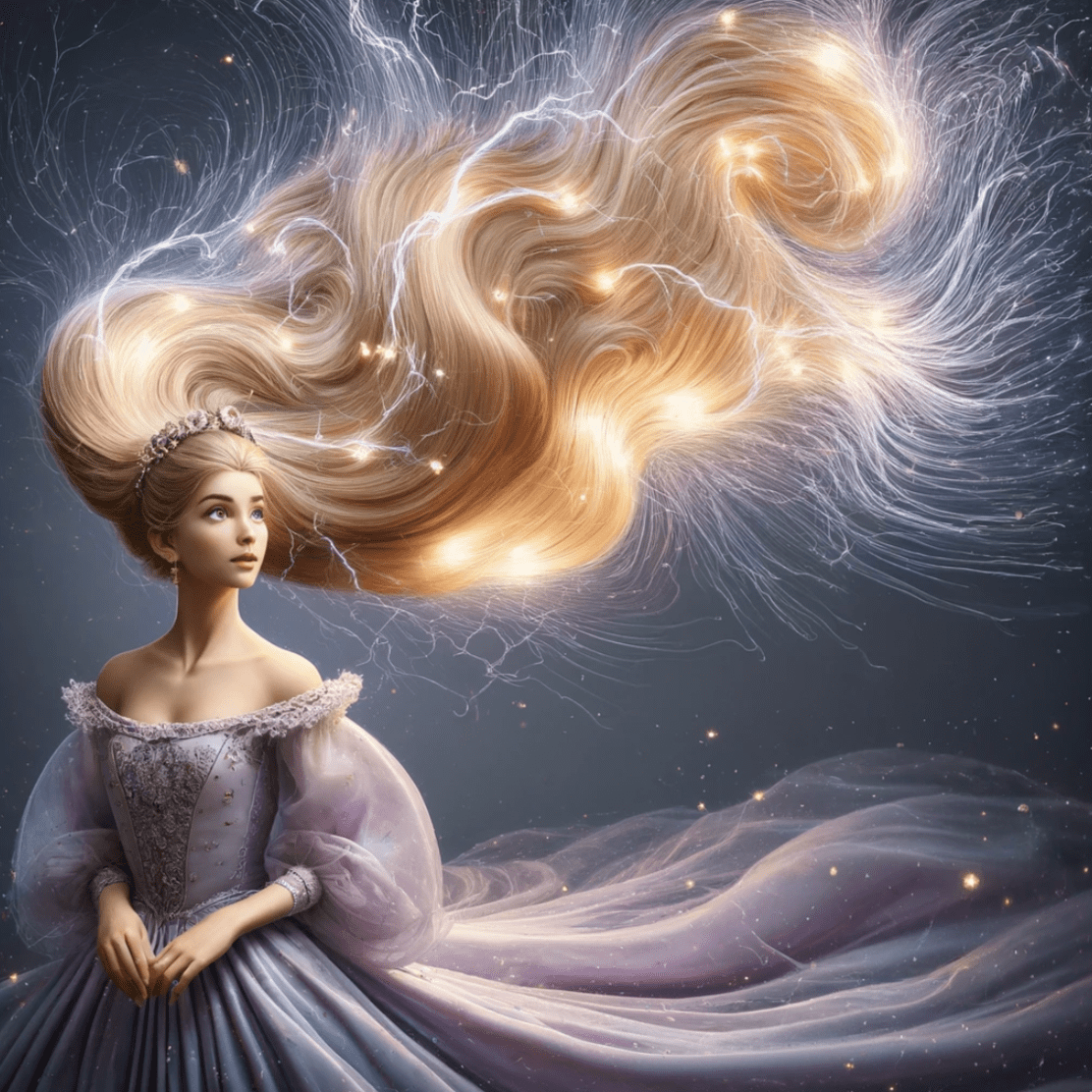}}       \\
    \textit{Exaggeration}     & In this type of editing, objects experience exaggerated transformations that cannot occur in the real world. & \raisebox{-0.8\totalheight}{\includegraphics[width=0.14\linewidth]{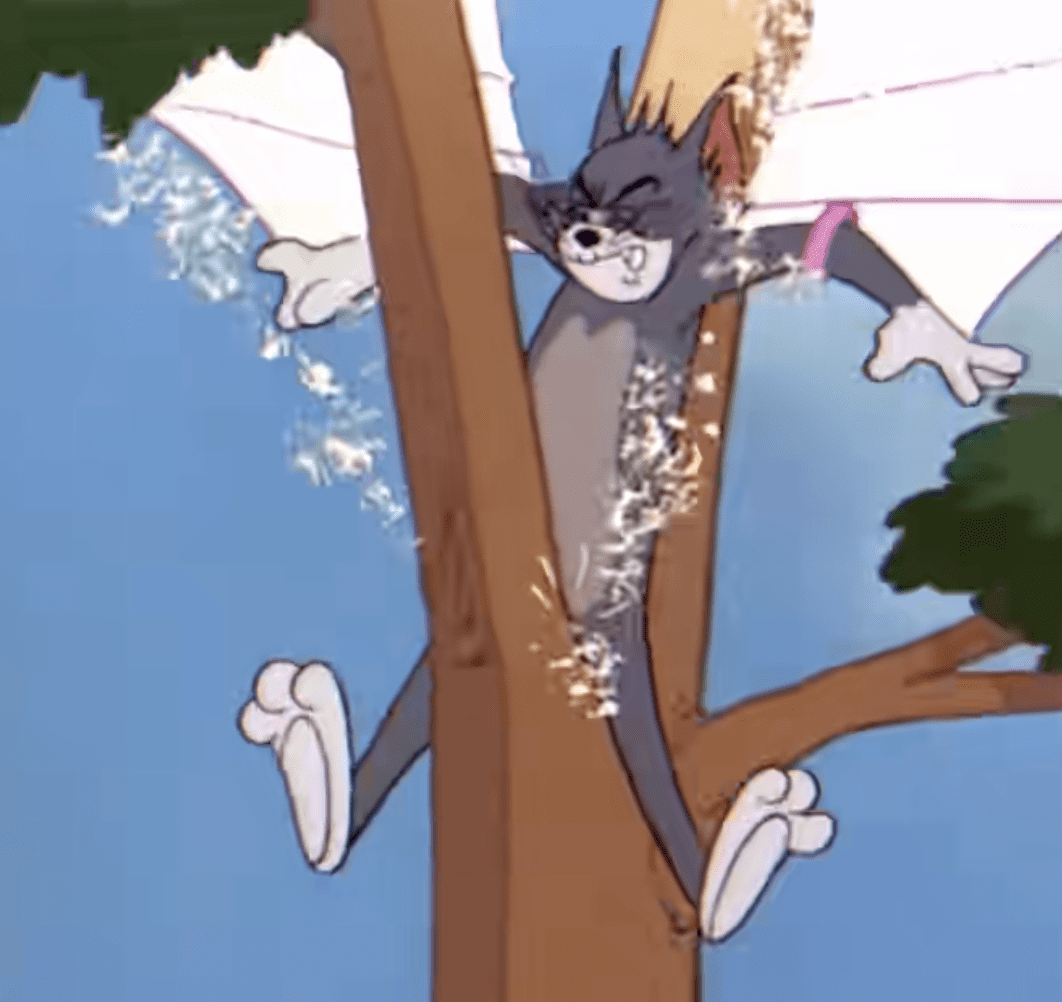}} & ``What would the cat look like after falling to the ground?'' & \raisebox{-0.8\totalheight}{\includegraphics[width=0.14\linewidth]{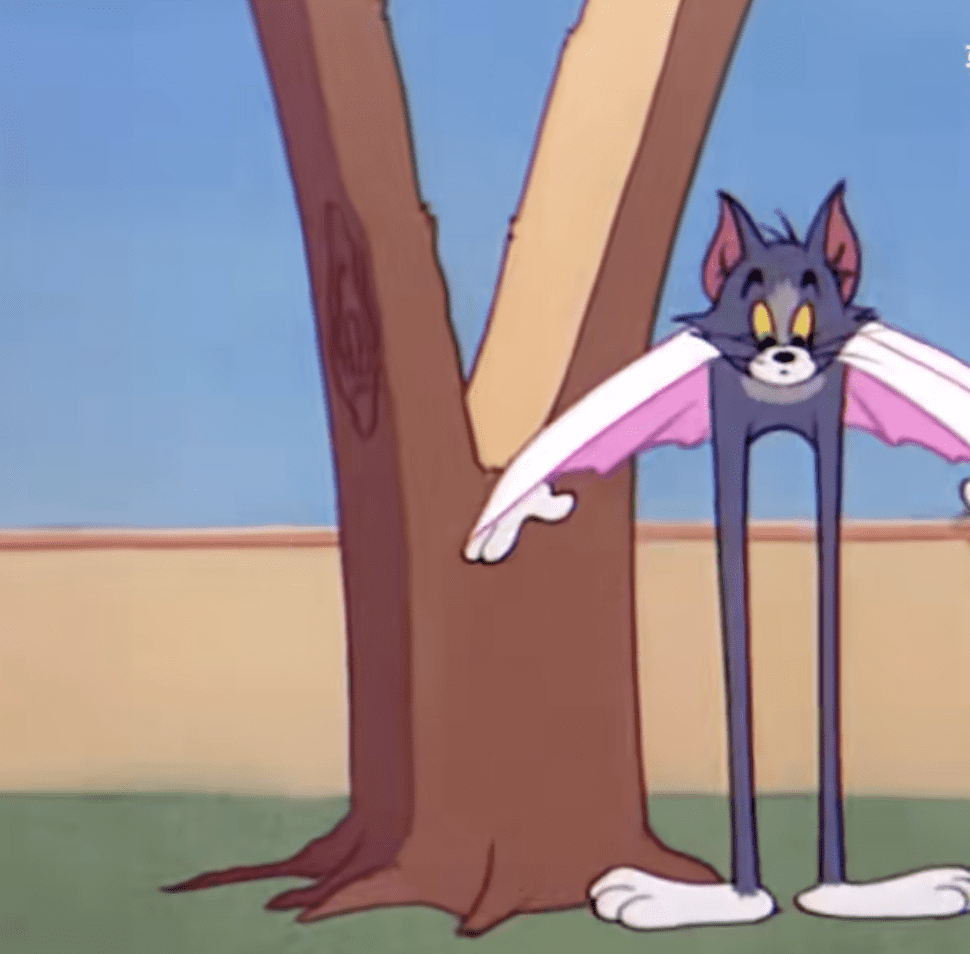}}       \\ \bottomrule
   \end{tabular}
   \vspace{-2mm}
\end{table}

\subsection{Generating A Dataset with World Instructions}
\label{subsec:data_pipeline}

\begin{figure}[t]
  \centering
  \includegraphics[width=0.98\linewidth]{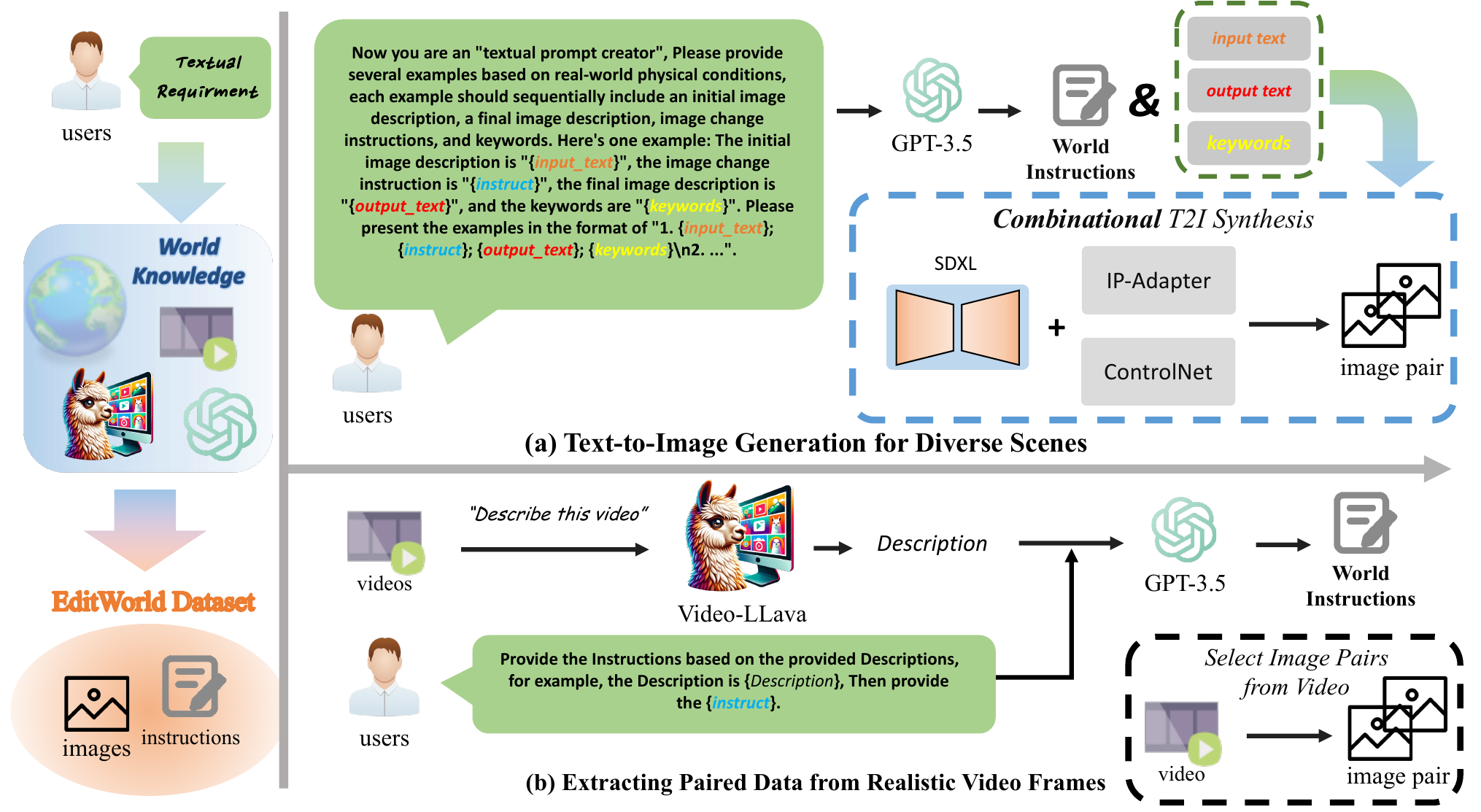}
  \caption{Generating a dataset of world-instructed image editing from two different branches.}
  \label{fig:data_pipeline}
\end{figure}

\subsubsection{Text-to-Image Generation for Diverse Scenarios}
In this part, we introduce how to leverage LLMs and text-to-image generation models to produce a large amount of input-instruction-output triplets. 
As shown in Fig.~\ref{fig:data_pipeline}(a), we design a template to prompt the GPT-3.5 to synthesize a series of textual quadruples, (prompt $y_{ori}$ of input image, instruction $y_{instr}$, prompt $y_{tar}$ of output image, and keywords $\{k_{1}$, $k_{2}$,..., $k_{n}\}$), which are used for later generation of input-output paired images.
Among them, keywords are used to denote the parts to be edited.

\paragraph{Combinational T2I Generation Procedure.} We incorporate a pretrained text-to-image diffusion denoiser $\epsilon_{\theta}$ of SDXL \cite{podell2024sdxl} 
to synthesize input image $I_{ori}$ and output image $I_{tar}$ based on the textual quadruple. Firstly, to localize the editing areas, we collect the cross-attention maps $M_{1}$, $M_{2}$,..., $M_{n}$ of $k_{1}$, $k_{2}$,..., $k_{n}$ during the generation of $I_{ori}$. We binarize each pixel $(m,n)$ of these attention masks using a threshold to create binary masks $M^{b}$. 
The binarization process is defined as follows:
\begin{align}
\begin{split}
    M^{b}_{i}(m,n) &= \begin{cases} 
    0, & \text{if } M_{i}(m,n) \leq 0.8125*\mathrm{mean}(M_{i}), \\
    1, & \text{if } M_{i}(m,n) > 0.8125*\mathrm{mean}(M_{i}),
    \end{cases}
\end{split}  \\
\begin{split}
    M^{b} &= M^{b}_{1} \cup M^{b}_{2} \cup ... \cup  M^{b}_{n}.
\end{split}
\end{align}

Then, to preserve the non-editing areas, we conduct image inpainting within the masked areas $M^{b}$ of $I_{ori}$ to synthesize $I_{tar}$ with the textual guidance $y_{tar}$. The $I_{ori}$ is encoded into latent space using the VAE encoder $\mathcal{E}$ of the LDM \cite{rombach2022high} $z_{ori} \sim \mathcal{E}(I_{ori})$, and the denoising-based inpainting process can be defined as:
\begin{equation}
\begin{aligned}
    z_{ori,t} &= \mathcal{N}(\sqrt{\alpha_t} z_{ori}, (1-\alpha_t)\mathbf{I}), 
    & z^{*}_{t} = z_{t} \cdot M^{b} + z_{ori,t} \cdot (1 - M^{b}), 
\end{aligned}
\label{eq:attnmask}
\end{equation}
where $z_{t}$ denotes the denoised latent. To further ensure the identity consistency between $I_{ori}$ and $I_{tar}$, we employ IP-Adapter \cite{ye2023ip} and ControlNet \cite{zhang2023adding} $\mathcal{F}_{ctrl}$ for additional refinement of $I_{tar}$. 
Specifically, IP-Adapter extracts visual feature $f_{ori}$ from the original image $I_{ori}$, which guides the refinement process to maintain the identity consistency between $I_{ori}$ and $I_{tar}$. Simultaneously, ControlNet utilizes the canny map $m_{canny}$ of $I_{tar}$ to guide the refinement, thereby preserving the structure of $I_{tar}$. The predicted noise of the refinement of $I_{tar}$ is denoted as:
\begin{equation}
\begin{aligned}
    \epsilon = \epsilon_{ip}(z_{t}, t, y_{tar}, f_{ori}, \mathcal{F}_{ctrl}(z_{t}, m_{canny})),
\end{aligned}
\label{eq:datafinalgen}
\end{equation}
where  $\epsilon_{ip}$ is the denoiser of IP-Adapter. Finally, we employ a MLLM as the discriminator, using semantic alignment, identity consistency, and image quality as criteria to select the most reasonable and high-quality generated image pairs as our training data. 
The types of world instructions that can be obtained with this text-to-image generation branch include \textit{Long-Term}, \textit{Physical-Trans}, \textit{Implicit-Logic}, \textit{Story-Type}, and \textit{Real-to-Virtual}. 

\paragraph{Expressiveness of Our Generation Procedure.} It is worth mentioning that, compared to InstructPix2Pix \cite{brooks2023instructpix2pix} which generates training data using simple cross-attention manipulation in Prompt-to-Prompt \cite{hertz2022prompt}, our text-to-image generation branch can handle more complex image changes by the combinational utilization of specific pretrained models. For example, in Prompt-to-Prompt, the input-output text modifications only allow simple word changes, such as "A snowman" to "A melted snowman". In contrast, our method allow a more complex change from "A snowman stands in a winter wonderland surrounded by glistening snowflakes" to "A melted snowman, leaving behind a puddle of water and a few remnants of coal and carrot on the ground". This means that the dataset generation of \method is more flexible and expressive, and the generated data can include more complex editing scenarios.

\subsubsection{Extracting Paired Data from Realistic Video Frames}
In addition to the rich generated data from text-to-image branch, we also extract paired data from a video dataset InternVid \cite{wang2024internvid} to add more realistic data into our dataset. More concretely, we select two video frames that have strong identity consistency but the greatest spatial/visual variances or dynamics caused by certain motions. The two frames usually are the starting and ending points of the video storyline. Then, as shown in Fig.~\ref{fig:data_pipeline}(b), we use a pretrained video-language model Video-LLava \cite{lin2023video} to obtain a description of this video storyline. Subsequently, we employ GPT-3.5 to transform this description into a world instruction according to a predefined format. In this way, the extracted first and last frames can serve as the input image and output image, respectively. Combined with the produced instructions, they can form a series of input-instruction-output triplets for augmenting our editing dataset. This branch of triplet examples extracted from video data include following types of world instructions: \textit{Spatial-Trans}, \textit{physical-Trans}, \textit{Story-Type}, and \textit{Exaggeration}.

\subsubsection{Statistics for The \method Dataset}

\begin{figure}[t]
  \centering
  \includegraphics[width=0.98\linewidth]{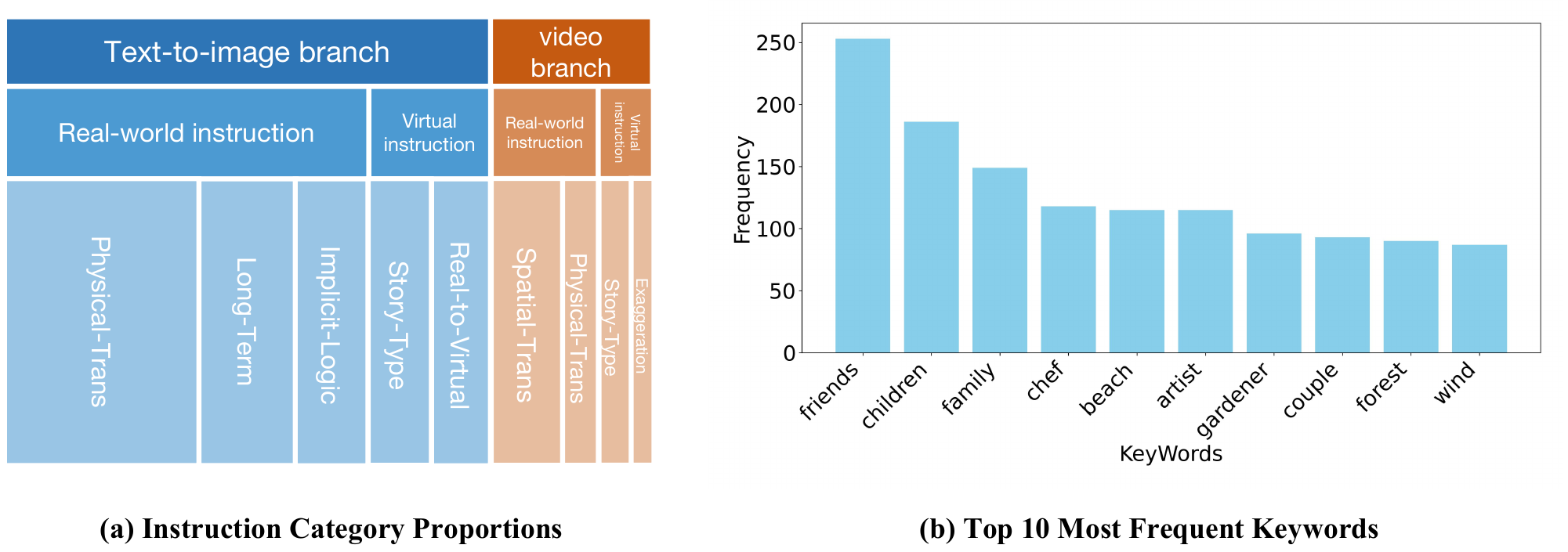}
  \caption{Statistics of our generated \method dataset.}
  \label{fig:data_state}
  \vspace{-3mm}
\end{figure}

To provide a more intuitive understanding, we visualize the details of our dataset. Fig.~\ref{fig:data_state}(a) shows the distribution of \method dataset based on different branches and categories of instructions. Since the quantities of distinct types of changes in world dynamics vary (for instance, changes in the physical world are more abundant than in the virtual world), the corresponding data volumes for various categories of world instruction in \method dataset differ accordingly. Additionally, to more comprehensively illustrate the contents of our dataset, Fig.~\ref{fig:data_state}(b) presents the 20 most frequently occurring keywords. Besides, some generated images of these two branches may have low resolution and lack reasonableness, we provide the specific recheck process in \cref{Sec:human_select}.

\begin{figure}[t]
  \centering
  \includegraphics[width=0.98\linewidth]{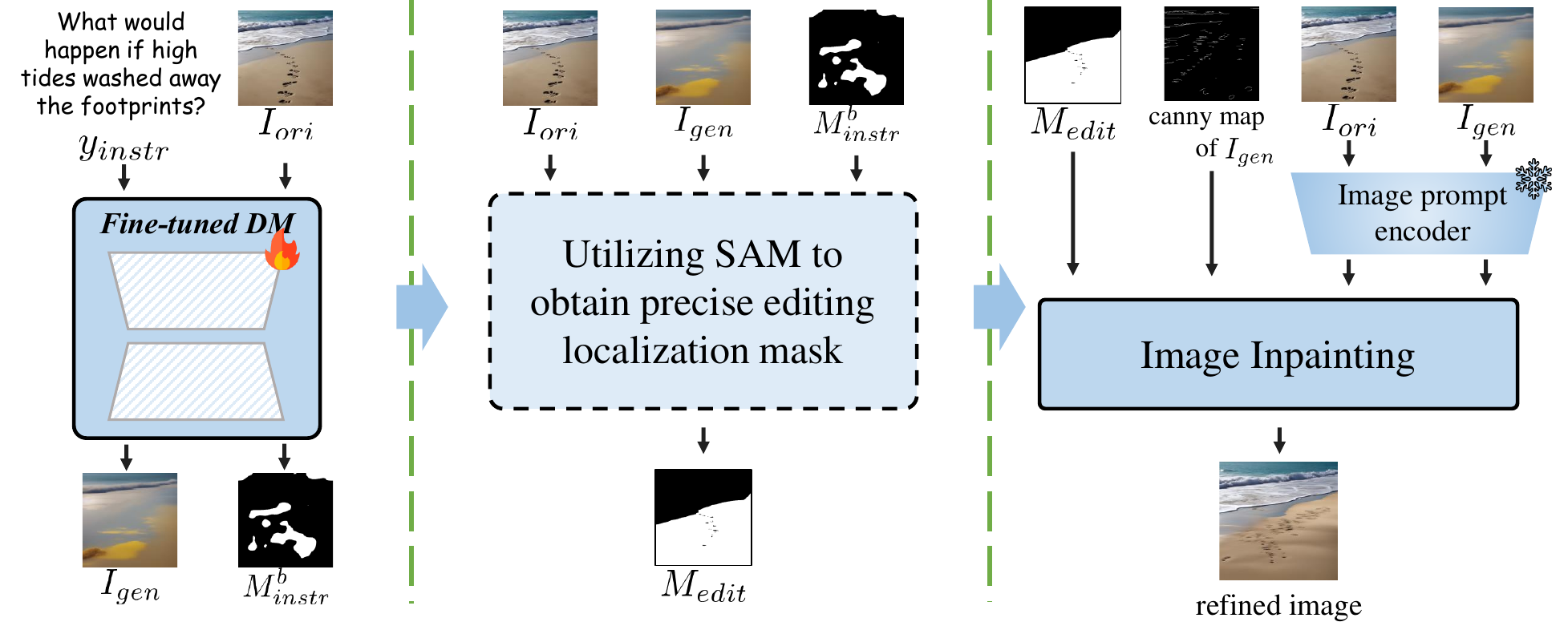}
  \caption{Illustrate our image manipulation method \textit{post-edit} for instruction-guided image editing.}
  \label{fig:framework}
\end{figure}

\subsection{Image Editing Model of \method}

We use the constructed dataset from the two branches to train an instruction-based image editing diffusion model (DM). The input conditions for the DM include the instruction $y_{instr}$ and the latent code $z_{ori}$ of input image $I_{ori}$. The supervision function for training the denoiser $\epsilon_{\theta}$ is as follows:
\begin{equation}
    \begin{aligned}
        L = \mathbb{E}_{z_t, t, z_{ori}, y_{instr}, \epsilon \sim \mathcal{N}(0,1)}[||\epsilon - \epsilon_{\theta}(z_t, z_{ori}, y_{instr}, t)||^2_2].
    \end{aligned}
\end{equation}

We define the generated result of the trained diffusion model as $I_{gen}$. Following the classifier-free guidance described in InstructPix2Pix \cite{brooks2023instructpix2pix}, we modify the noise estimation based on the $I_{ori}$ and the $y_{instr}$ as follows:
\begin{equation}
    \begin{aligned}
        \hat{\epsilon}_{\theta}(z_t, I_{ori}, y_{instr}) =& \ \epsilon_{\theta}(z_t, \varnothing, \varnothing) + s_I \cdot (\epsilon_{\theta}(z_t, I_{ori}, \varnothing) - \epsilon_{\theta}(z_t, \varnothing, \varnothing)) \\
        &+ s_T \cdot (\epsilon_{\theta}(z_t, I_{ori}, y_{instr}) - \epsilon_{\theta}(z_t, I_{ori}, \varnothing)),
    \end{aligned}
\end{equation}

where $s_I$ and $s_T$ denote the guidance scales. 

\paragraph{Post-Edit Method} 
The fine-tuned DM has been able to make effective world-instructed image editing.
To further refine the generated results, and maintain the non-edited areas of the input image as much as possible, as shown in Fig.~\ref{fig:framework}, we propose a new image manipulation method \textit{post-edit} to optimize the editing process. 
Specifically, we extract a binary attention mask $M^{b}_{instr}$ from $\epsilon_{\theta}(z_t, I_{ori}, \varnothing)$ during the inference process, providing approximate localization of editing position. 
Then, we use SAM \cite{kirillov2023segment} to segment both $I_{gen}$ and $I_{ori}$, and calculate the precise masks $M_{gen}$ and $M_{ori}$ which have the maximum overlaps with $M^{b}_{instr}$. Subsequently, we define the union of $M_{gen}$ and $M_{ori}$ as the mask $M_{edit}$, which enhances the precision of the editing localization. While $I_{gen}$ has already offered a reasonable editing results, putting $M_{edit}$, canny map of $I_{gen}$, and visual features of $I_{ori}$ and $I_{ori}$ into an image inpainting process, our method can further boost the quality of $I_{gen}$ while preserving the non-edited areas well.


\section{Experiment}

\subsection{Implementation Details}
\label{Sec:impl_detail}

\paragraph{Evaluation}
We randomly selected 300 data triples from the text-to-image branch and 200 data samples from the videos branch for testing. The remaining data were utilized to train the image editing model of \method. Each test data triple from the testing dataset includes an input image, instruction, and an output image description. We employ the CLIP score as a metric to assess whether the generated editing results align with the output semantics, thus judging the effectiveness of the method in world-instructed editing. 
Additionally, due to the complexity of the world-instructed editing task, the CLIP score may be insufficient to accurately evaluate the logical accuracy of the image editing in some cases. Therefore, we introduce a new metric, termed as \textit{MLLM score}, which employs multimodal large language models to assess the instruction-following ability of various methods in world-instructed editing task. Comprehensive evaluation details and a full list of validation results regarding MLLM score can be found in \cref{app-mllm-score}.

\paragraph{Training Details}
Data generation and model training of \method are completed on 4$\times$80GB NVIDIA A100 GPUs. The model, initialized with the weights of InstructPix2Pix \cite{brooks2023instructpix2pix}, is fine-tuned on our proposed dataset for a total of 100 epochs. We use a batch size of 64 and an image resolution of 512×512. The training employs the Adam optimizer \cite{kingma2014adam} with a learning rate of 5e-6.

\subsection{Quantitative Evaluation}
To more accurately assess the editing capabilities of various models across different scenarios in world-instructed image editing, this study tests image editing abilities under different data types, categorized by seven types of instructions and two data branches. We compare our \method with five SOTA methods \cite{parmar2023zero, couairon2022diffedit, brooks2023instructpix2pix, zhang2023magicbrush, fu2024guiding} under different types of instruction in Table~\ref{tab:exp_t2i_clip}.

\paragraph{Comparing Editing Methods Between Text-to-Image and Video Branches}
Due to the different motivations for data acquisition in our text-to-image and video branches, we evaluate the performance of editing methods on the testing datasets of these branches separately. Our method consistently outperforms others in terms of CLIP scores and MLLM score metrics in two datasets, demonstrating superior editing capabilities and stability in world-instructed editing tasks. Remarkably, our method maintains excellent performance without post-editing, which highlights the effectiveness of our curated dataset. The comparison results reveal distinct variations in the performance enhancements throughout two testing datasets: within the text-to-image branch's testing dataset, our method generally signally outperform other methods, indicating that the sophisticated editing scenarios remain a challenge for traditional editing methods. Meanwhile, in the video branch's testing dataset, the improvement of our method is less significant than in the text-to-image branch's testing dataset. We attribute this difference to the fact that extracted frames from videos often contain complex scenes where the main subject is not prominent. Consequently, in the video branch, it is challenging to achieve ideal edits.
Besides, as shown in Table~\ref{fig:Qualitative}, while InstructPix2Pix achieves a higher CLIP score than other baseline methods, it indicates that although InstructPix2Pix responds effectively to the instructions, it still struggles to maintain the visual quality of the input images and realize desirable image editing in world-instructed editing.

\begin{table}[t]
\caption{Quantitative comparison of CLIP score and MLLM score. IP2P: InstructPix2Pix \cite{brooks2023instructpix2pix}; MB: MagicBrush \cite{zhang2023magicbrush}. Bold results are the best, and underlined ones are the second best.} 
\label{tab:exp_t2i_clip}
\centering
\setlength{\tabcolsep}{3.5pt}
\renewcommand{\arraystretch}{1.4}
\begin{tabular}{lccccccc}
\toprule
\multirow{2}{*}{Category} & \multicolumn{2}{c}{Text-based Method}                                                                                      & \multicolumn{5}{c}{Instruction-based Method}                                                                 \\ \cmidrule(r){2-3} \cmidrule(r){4-8} 
                        & \multicolumn{1}{c}{Pix2Pix-Zero} & \multicolumn{1}{c}{DiffEdit}   & \multicolumn{1}{c}{IP2P} & \multicolumn{1}{c}{MB} & \multicolumn{1}{c}{MGIE} & \multicolumn{1}{c}{\textbf{\method}}    & \multicolumn{1}{c}{\textbf{w/o post-edit}}              \\ \hline
\multicolumn{8}{c}{{CLIP Score of Text-to-image Branch}} \\ \hline
\textit{Long-Term}          & 0.1831  & 0.1952  & 0.2140  & 0.1870  & 0.1863  & \underline{0.2244}          & \textbf{0.2294} \\
\textit{Physical-Trans}     & 0.2266  & 0.2283  & 0.2186  & 0.2101  & 0.2286  & \underline{0.2385}          & \textbf{0.2467} \\
\textit{Implicit-Logic}     & 0.2356  & 0.2307  & 0.2390  & 0.2432  & 0.2350  & \textbf{0.2542}          & \underline{0.2440} \\    
\textit{Story-Type}         & 0.2111  & 0.1980  & 0.2063  & 0.2070  & 0.2084  & \textbf{0.2534}          & \underline{0.2354} \\ 
\textit{Real-to-Virtual}    & 0.2294  & 0.2419  & 0.2285  & 0.2344  & 0.2305  & \textbf{0.2524}          & \underline{0.2435} \\
\hline
\multicolumn{8}{c}{{CLIP Score of Video Branch}} \\ \hline
\textit{Spatial-Trans}      & 0.2102  & 0.2058  & 0.2175  & 0.1997  & 0.2157  & \textbf{0.2420}          & \underline{0.2286} \\
\textit{Physical-Trans}     & 0.1899  & {0.2407}  & 0.2315  & 0.2278  & 0.2277  & \underline{0.2467}          & \textbf{0.2483} \\
\textit{Story-Type}         & 0.1724  & 0.2114  & {0.2318}  & 0.2262  & 0.2155  & \underline{0.2365}          & \textbf{0.2399} \\ 
\textit{Exaggeration}       & 0.2164  & 0.2275  & {0.2416}  & 0.2328  & 0.2208  & \textbf{0.2443}          & \underline{0.2433} \\ 
\hline
\multicolumn{8}{c}{{MLLM Score of Both Branches}} \\ \hline
Text-to-image               & 0.8465  & 0.8517  & 0.8763  & 0.8455  & 0.8173  & \underline{0.8958}        & \textbf{0.9060} \\ 
Video                       & 0.8706  & 0.94  & 0.9493  & 0.9715  & 0.9474  & \textbf{0.9920}        & \underline{0.9891}\\
\bottomrule
\end{tabular}
\end{table}

\paragraph{Evaluation of Editing Methods in Various Instruction Categories}
To investigate the editing capabilities of various methods across different scenarios in world-instructed editing tasks, we evaluated their performance across seven distinct categories of instructions. Our method demonstrated superior performance in all categories. Particularly in categories such as \textit{Long-Term}, \textit{Physical-Trans}, \textit{Spatial-Trans}, \textit{Story-Type}, and \textit{Exaggeration}, our method excels as these involve significant differences between the input and output images while maintaining a relevant connection between them. The models used for traditional editing, lacking the knowledge of world dynamics, fail to generate plausible results in these scenarios, thus highlighting the effectiveness of our approach.
For categories \textit{Implicit-Logic} and \textit{Real-to-Virtual}, where instructions are complex or abstract and the changes between the input and output images are subtle, some changes are akin to those encountered in traditional editing tasks. Existing text-based editing methods \cite{parmar2023zero, couairon2022diffedit} perform well since they are directly provided with output text, simplifying their task. Conversely, instruction-based methods like InstructPix2Pix, which are trained with datasets featuring instructions generated by LLMs, excel in managing these intricate instructions. Despite this, our method consistently outperforms these baseline methods in these two categories.

\subsection{Qualitative Analysis}

\begin{figure}[t]
  \centering
  \includegraphics[width=1.0\linewidth]{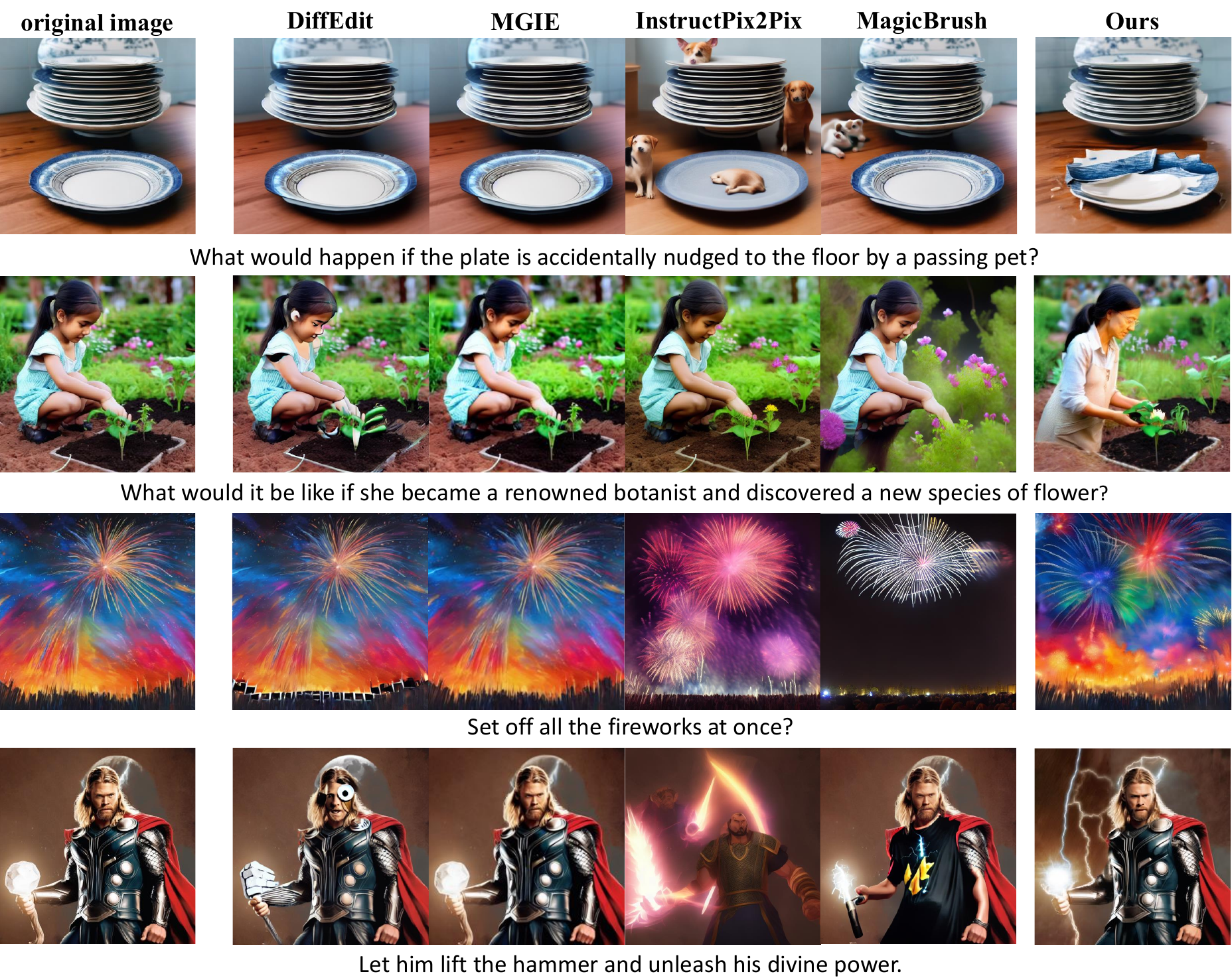}
  \caption{Qualitative comparison of world-instructed image editing.}
  \label{fig:Qualitative}
  \vspace{-4mm}
\end{figure}

\begin{figure}[t]
  \centering
  \includegraphics[width=0.98\linewidth]{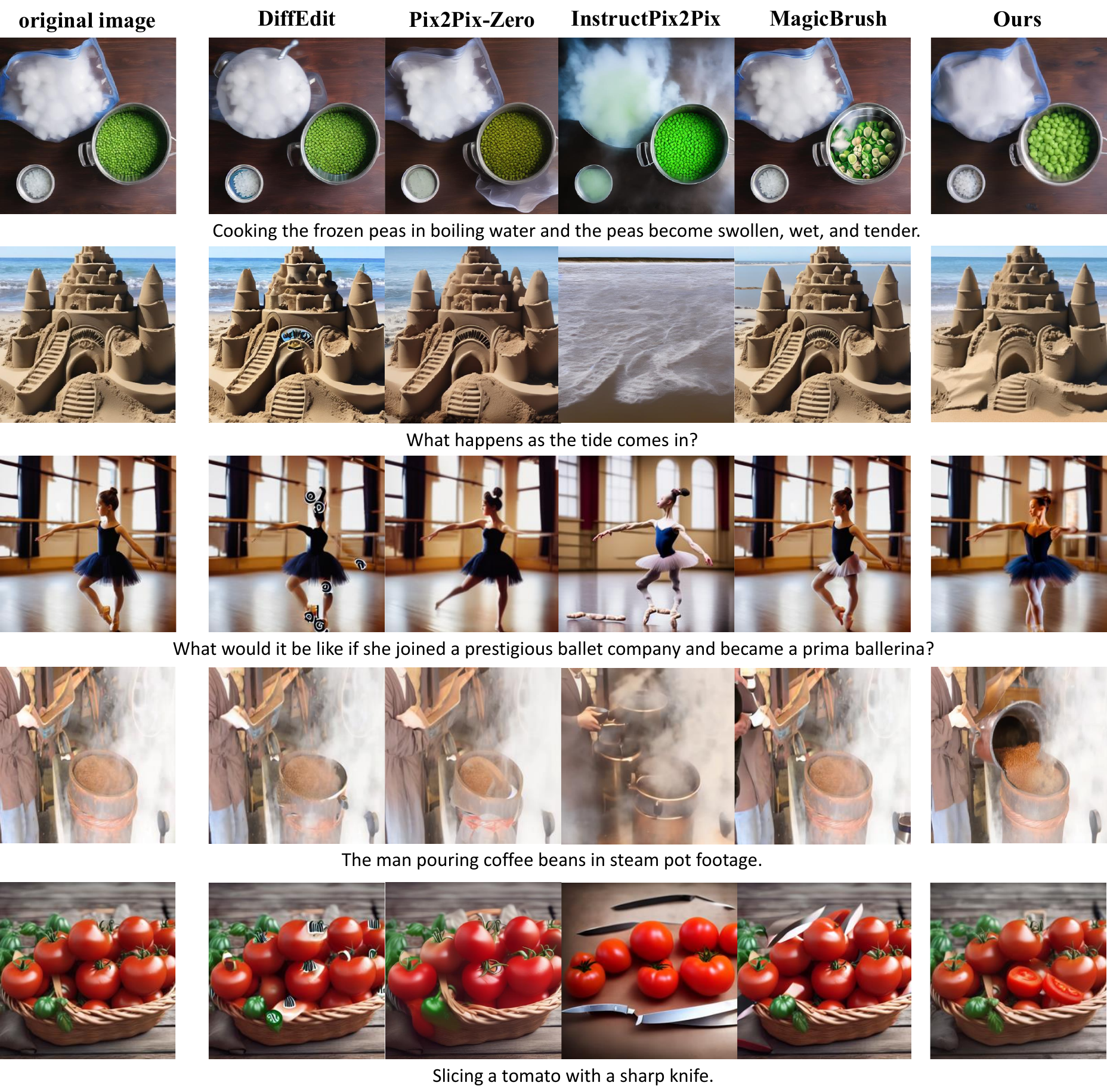}
  \caption{Qualitative comparison of world-instructed image editing.}
  \label{fig:Qualitative_supp}
\end{figure}

To demonstrate the editing capabilities of the \method model, we randomly selected several examples from different instructions for visualization, as shown in \cref{fig:Qualitative,fig:Qualitative_supp}. Previous editing methods were generally unable to correctly edit images based on the given text or instructions, resulting in either no changes to the image or significant alterations that did not meet the editing requirements. In contrast, our method can effectively edit images according to the given instructions, producing high-quality results. 
Notably, while DiffEdit \cite{couairon2022diffedit} and MGIE \cite{fu2024guiding} maintain the quality of the input images, their visualizations reveal a lack of accurate reflection of the editing instructions.
We provide more qualitative results in \cref{app-qualitative}.

\subsection{Ablation Study}

To validate that our proposed \textit{post-edit} method effectively preserves non-edited areas while maintaining high-quality editing results, and performs well in traditional editing tasks, we conducted an ablation study. We selected 50 traditional editing test datasets, including 18 input-instruction-output triples for 'add', 25 triples for 'change', and 7 triples for 'remove'. As presented in Table~\ref{tab:abla_lpips}, the LPIPS metric \cite{zhang2018unreasonable} significantly improves with \textit{post-edit}, whereas the CLIP score remains consistent, This demonstrates that our \textit{post-edit} can preserve the non-edited areas without degrading the editing quality. At the same time, our \method achieves comparable results to \cite{brooks2023instructpix2pix, zhang2023magicbrush} on the CLIP score, confirming its effectiveness in traditional editing tasks.

\begin{table}[t]
\caption{Ablation study, using the LPIPS distance between the generated edited results and the input image, tests the effectiveness of post-edit in three methods: instructPix2Pix (IP2P), MagicBrush (MB), and our \method.}
\label{tab:abla_lpips}
\centering
\setlength{\tabcolsep}{4pt}
\renewcommand{\arraystretch}{1.2}
\begin{tabular}{l|cccccc}
\toprule
Metrics                  & IP2P & IP2P w/ post-edit & MB & MB w/ post-edit  & Ours   & Ours w/o post-edit      \\ \hline
LPIPS                    & 0.3929  & 0.3465  & 0.2397  & 0.2358  & 0.3082  & 0.3778     \\
CLIP score               & 0.2317  & 0.2332  & 0.2383  & 0.2357  & 0.2344  & 0.2334     \\
\bottomrule
\end{tabular}
\end{table}

\section{Conclusion}
\vspace{-2mm}

We propose a new image editing task called world-instructed image editing, which uses different scenarios in the real and virtual worlds to provide editing instructions for image editing. We classify, define, and exemplify these instructions, and use GPT, large-scale multimodal models, and text-to-image generation models to obtain image editing data from a large number of videos and texts. Using this data, we construct quantitative evaluation metrics for the world-instructed image editing task, and use the collected data to train and improve the image editing model, achieving state-of-the-art (SOTA) performance in this new world-instructed image editing task.

{\small
\bibliographystyle{ieeetr}
\bibliography{main}
}


\newpage

\appendix

\section*{\LARGE\textbf Appendix}

\section{\method Dataset}

\subsection{Human Selection}
\label{Sec:human_select}

Utilizing text-to-image generation and video branches to obtain our dataset, some data triples may lack reasonableness despite the rich content and diverse scenarios. To further enhance the data quality, we employ some workers to recheck our curated dataset. Their specific tasks include: 1) For triples where the text-to-image generation results are unsatisfactory, adjust the prompt to improve the quality of the synthesis results; 2) Remove some editing data from the video branch whose images are unclear and blurry; 3) Revise unreasonable instructions in the data triples obtained from videos. At the same time, we discover that for some prompts, SDXL cannot generate desirable results, such as "balloon bursting" and "sandcastle being washed away by the sea". To address this, we adopt DALL·E~3 to generate and select higher quality images to replace the triples generated by SDXL.



\subsection{MLLM Score}
\label{app-mllm-score}
\begin{table}[h]
\caption{Quantitative comparison of MLLM score. IP2P: InstructPix2Pix \cite{brooks2023instructpix2pix}; MB: MagicBrush \cite{zhang2023magicbrush}.}
\label{tab:exp_t2i_mllmscore}
\centering
\setlength{\tabcolsep}{3.5pt}
\renewcommand{\arraystretch}{1.4}
\begin{tabular}{lcccccccc}
\toprule
\multirow{2}{*}{Category} & \multicolumn{2}{c}{Text-based Editing}                                                                                      & \multicolumn{5}{c}{Instruction-based Editing}                                                                 \\ \cmidrule(r){2-3} \cmidrule(r){4-8} 
                        & \multicolumn{1}{c}{Pix2Pix-Zero} & \multicolumn{1}{c}{DiffEdit}   & \multicolumn{1}{c}{IP2P} & \multicolumn{1}{c}{MB} & \multicolumn{1}{c}{MGIE} & \multicolumn{1}{c}{\textbf{\method}}    & \multicolumn{1}{c}{\textbf{w/o post-edit}}      \\ \hline
\multicolumn{8}{c}{\textbf{Text-to-image Generation Branch}} \\ \hline
\textit{Long-Term}          & 0.9871  & 0.9846  & 0.9846  & \underline{0.9874}  & 0.9743  & \textbf{1.0}           & 0.9871  \\   
\textit{Physical-Trans}     & 0.8381  & 0.8476  & 0.8571  & 0.8452  & 0.8357  & \textbf{0.9095}        & \underline{0.9023}  \\    
\textit{Implicit-Logic}     & 0.8047  & 0.8323  & 0.8809  & 0.8238  & 0.7857  & \underline{0.8810}        & \textbf{0.9048}  \\   
\textit{Story-Type}         & 0.7652  & 0.7857  & {0.8047}  & 0.7714  & 0.6619  & \underline{0.8429}        & \textbf{0.8762}  \\ 
\textit{Real-to-Virtual}    & 0.8375  & 0.8083  & {0.8542}  & 0.8  & 0.8292  & \textbf{0.8658}          & \underline{0.8583}  \\
\hline
\multicolumn{8}{c}{\textbf{Video Extraction Branch}} \\ \hline
\textit{Spatial-Trans}      & 0.9271  & 0.9152  & {0.9878}  & 0.9608  & 0.983  & \underline{0.997}  & \textbf{1.0}    \\
\textit{Physical-Trans}     & 0.8435  & 0.9791  & \textbf{0.9875}  & 0.9792  & 0.9458  & \textbf{0.9875}   & 0.9846   \\
\textit{Story-Type}         & 0.8175  & 0.9382  & 0.9448  & {0.989}  & 0.961  & \textbf{1.0}     & \underline{0.994}  \\ 
\textit{Exaggeration}       & 0.8944  & 0.9277  & 0.8774  & {0.957}  & 0.9  & \textbf{0.9834}    & \underline{0.9778}  \\ 
\bottomrule
\end{tabular}
\end{table}

Due to the limitations of the CLIP score in demonstrating the capability of world-instructed editing, in this section, we introduce the use of the MLLM Video-LLava \cite{lin2023video} and propose a new metric, MLLM score, to validate the effectiveness of world-instructed editing. Specifically, we present edited images, input and output texts, and instructions for the Video-LLava model. The specific prompts provided for Video-LLava for scoring the edited images are as follows: \zbh{"The input description <input text>, the editing instruction <instruction>, and the output description <output text>. Please evaluate if the given edited image has been successfully edited. if you think editing is successful, just give me 1, else if you think editing fails, just give me 0"}. As shown in Table~\ref{tab:exp_t2i_mllmscore}, our method continues to perform best under the MLLM score. Simultaneously, the performance of InstructPix2Pix \cite{brooks2023instructpix2pix} in the MLLM score is significantly worse than that in the CLIP score. Qualitative comparisons suggest that InstructPix2Pix often produces unreasonable results, further validating the superiority of the MLLM score over the CLIP score. Additionally, while the MLLM score is a reasonable metric for world-instructed editing, it struggles to differentiate subtle variations. Therefore, if a more effective metric is desired to evaluate the outcomes of world-instructed editing, it is necessary for the MLLM to more accurately distinguish differences between images or videos."

\section{More Qualitative Results}
\label{app-qualitative}

\begin{figure}[h]
  \centering
  \includegraphics[width=0.98\linewidth]{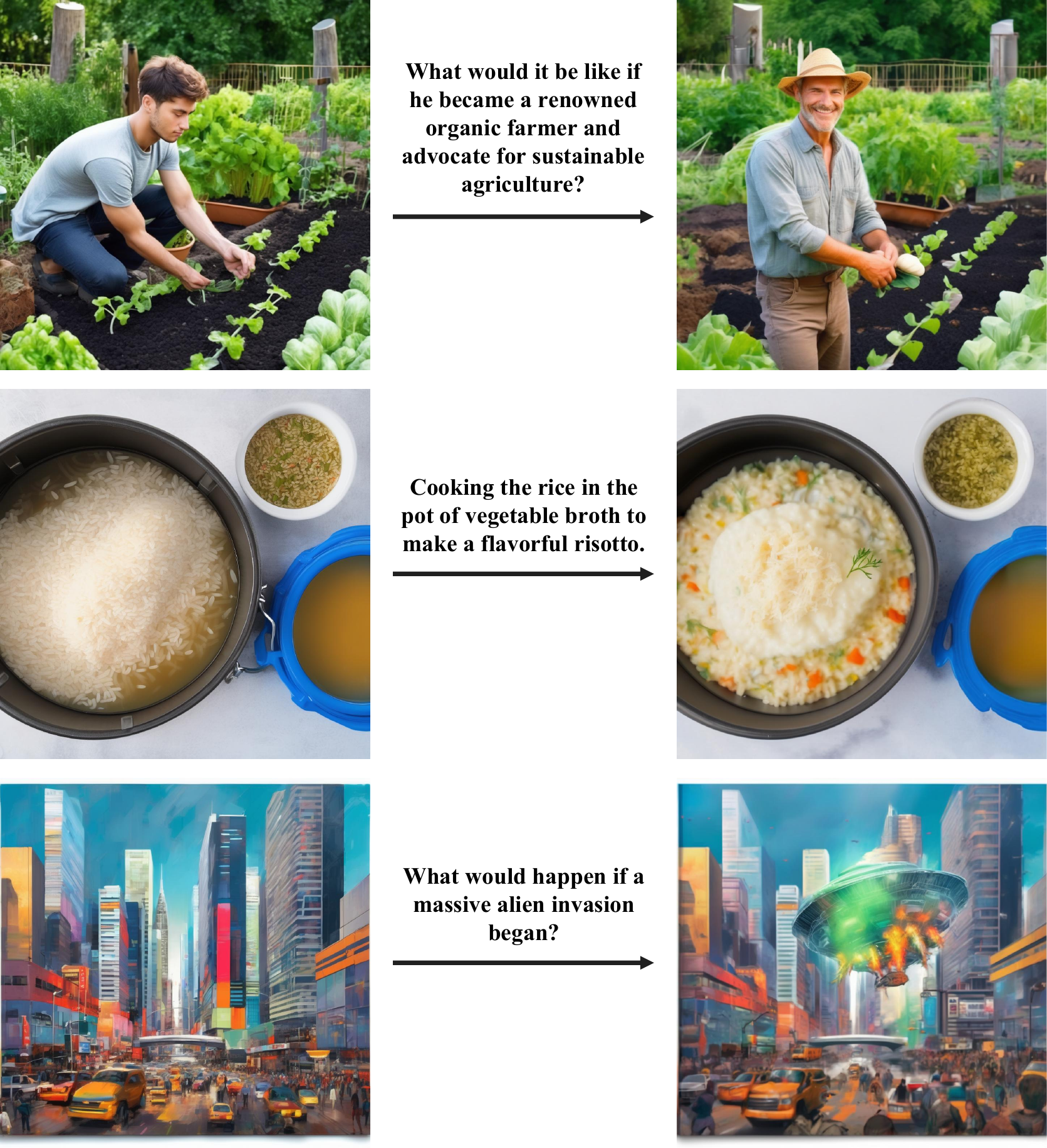}
  \caption{More editing results of \method}
  \label{fig:more_edit_supp}
\end{figure}

\begin{figure}[h]
  \centering
  \includegraphics[width=0.98\linewidth]{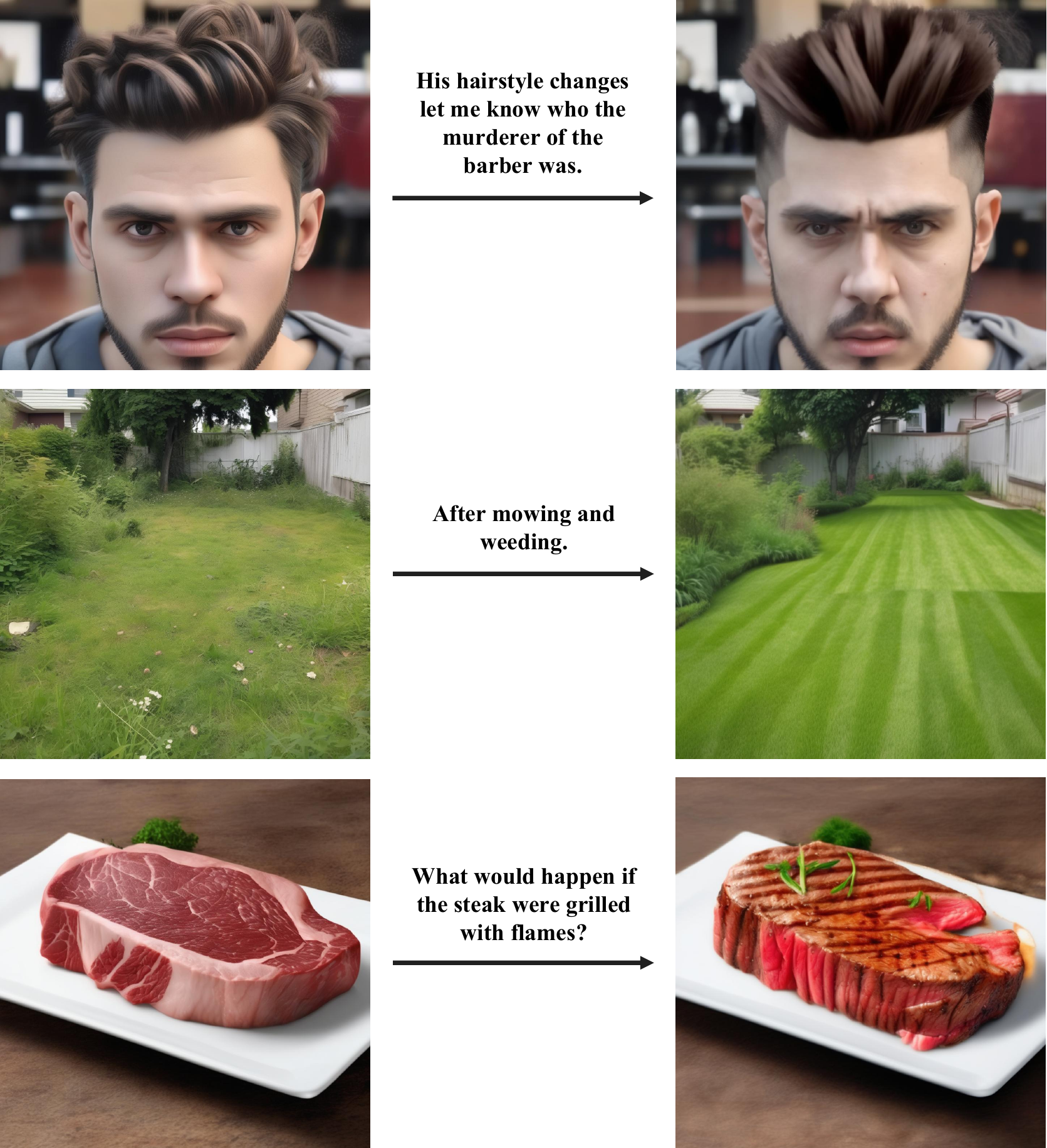}
  \caption{More editing results of \method}
  \label{fig:more_edit_supp_2}
\end{figure}

To further demonstrate the generation quality of \method, we present some high-resolution editing results in Fig.~\ref{fig:more_edit_supp} and Fig.~\ref{fig:more_edit_supp_2}.

\section{Limitation and Discussion}
\label{Sec:limit_discuss}

\paragraph{Limitation.}
Although our proposed world-instructed image editing task is of great practical significance, the scenarios in the real and virtual worlds are too complex and rich, and the amount of data we have collected is far from sufficient for true world editing. While the integration of MLLM with our method realizes world-instructed editing in a broader range of scenarios, efficient editing still requires a substantial amount of high-quality image editing training data. In addition, our data lacks a large number of precise editing examples. For instance, in a picture with eight candles arranged, given the instruction "blow out the third candle", accurate implementation of such image editing is almost non-existent in the data we have collected. Therefore, precise world-instructed image editing in complex pictures is still a challenging problem. Hence, our future work will focus on further enriching the \method dataset and increasing the number of precise editing data samples.

\paragraph{Discussion.}
Current large-scale multimodal models, such as LLava and GPT-4V, have demonstrated efficacy in answering questions based on content derived from videos and images.  However, these models encounter difficulties when tasked with distinguishing differences between images or videos.  For instance, they struggle to describe the differences or dynamics between input and output images within the dataset we have proposed.  This issue primarily stems from the need for large-scale data to analyze differences between images or videos. The task and data proposed in \method can contribute to addressing the challenge of multimodal difference recognition. In summary, both world-instructed image editing task and the recognition of differences in images or videos are indispensable parts in the pursuit of Artificial General Intelligence.

\section{Social Impact}
\label{Sec:impact}

Our \method does not have direct negative impacts on society, we must still be cautious about the potential misuse of world-instructed editing for illicit purposes.



\end{document}